\documentclass{article}

\usepackage[T1]{fontenc}
\usepackage[utf8]{inputenc}
\usepackage{microtype}
\usepackage{graphicx}
\usepackage{subcaption}
\usepackage{booktabs}
\usepackage{multirow}
\usepackage{amsmath}
\usepackage{amssymb}
\usepackage{mathtools}
\usepackage{xcolor}
\usepackage{upquote}
\usepackage[most]{tcolorbox}
\usepackage{tikz}
\usepackage{hyperref}

\usepackage[preprint]{icml2026}

\usepackage[capitalize,noabbrev]{cleveref}
\usetikzlibrary{arrows.meta,positioning}

\tcbset{
  json listing style/.style={
  enhanced,
  listing only,
  colback=blue!2,
  colframe=blue!45!black,
  colbacktitle=blue!45!black,
  coltitle=white,
  fonttitle=\bfseries\footnotesize,
  arc=2mm,
  boxrule=0.6pt,
  left=1mm,
  right=1mm,
  top=1mm,
  bottom=1mm,
  listing options={
    basicstyle=\ttfamily\tiny,
    breaklines=true,
    columns=fullflexible,
    keepspaces=true,
    showstringspaces=false,
    upquote=true
  }
  }
}

\tikzset{
  pipelinebox/.style={
    draw=blue!45!black,
    fill=blue!2,
    rounded corners=2mm,
    align=center,
    minimum height=9mm,
    font=\small
  },
  pipelinearrow/.style={-{Latex[length=2mm]}, thick, draw=blue!45!black}
}

\newtcblisting{jsonbox}{json listing style, title={Target JSON}}
\newtcblisting{schemajsonbox}{json listing style, title={Target Schema}, width=0.55\textwidth, center}
\newtcblisting{eventjsonbox}{json listing style, title={Event Forms}, width=0.70\textwidth, center}
\newtcblisting{promptbox}[1][]{json listing style, title={Prompt}, width=0.95\textwidth, center,#1}

\newcommand{\scoreup}{\ensuremath{\uparrow}}
\newcommand{\scoredown}{\ensuremath{\downarrow}}
\newcommand{\bestscore}[1]{\underline{\textbf{#1}}}
\newcommand{\scoreci}[3]{\begin{tabular}[c]{@{}c@{}}#1\\[-0.35ex]{\tiny[#2,#3]}\end{tabular}}
\newcommand{\bestci}[3]{\begin{tabular}[c]{@{}c@{}}\underline{\textbf{#1}}\\[-0.35ex]{\tiny[#2,#3]}\end{tabular}}

\icmltitlerunning{\smash{Geo-Strat-RL}}

\begin{document}

\twocolumn[
  \icmltitle{Geo-Strat-RL: Learning Geological Event Reasoning from Verifiable Tasks}

  \begin{icmlauthorlist}
    \icmlauthor{Lukas Mosser}{project}
  \end{icmlauthorlist}

  \icmlaffiliation{project}{Aker BP ASA}
  \icmlcorrespondingauthor{Lukas Mosser}{Lukas.mosser@akerbp.com}

  \icmlkeywords{vision-language models, reinforcement learning, synthetic benchmarks, geology, sequence stratigraphy}

  \vskip 0.3in
\begin{abstract}
To evaluate whether vision-language models can reason about geological histories, it is necessary to construct observations for which the underlying process history is known.
Furthermore, reasoning over geological histories is not just a question of recognizing visual patterns, but also of understanding temporal and structural relationships that may be only indirectly visible or highly ambiguous.
When ground-truth event histories are not uniquely identifiable or are unavailable, it remains an open challenge to teach models capable of visual reasoning to produce valid geological reconstructions that are consistent with both observed evidence and geological principles.
We therefore investigate whether defining a verifiable geological reasoning task can improve geological event reconstruction across observation domains through reinforcement learning with verifiable rewards (RLVR).
To this end, we present Geo-Strat-RL, a synthetic environment that generates stratigraphic observations and compact visible-evidence event histories.
The environment combines a geological generator with an executable verifier that scores chronology, event identity, deposition, and structural relationships.
We show that RLVR improves geological reconstruction in vision-language models (VLMs), increasing geological content scores on held-out stratigraphic diagrams.
We further evaluate the same held-out geological histories in a synthetic seismic observation domain by converting the generated scenes into acoustic-impedance-derived amplitude sections.
In this controlled paired-renderer setting, we present evidence that geological reasoning learned from stratigraphic diagram-domain RLVR training transfers to synthetic seismic representations without seismic-specific training examples, supporting the hypothesis that RLVR can teach reusable geological reasoning concepts across related observation formats.
\end{abstract}
]

\printAffiliationsAndNotice{}

\section{Introduction}
\label{sec:introduction}

Machine learning has a long history in geoscience, spanning remote sensing, seismic interpretation, lithology classification, document-scale knowledge extraction, and many more applications \citep{bergen2019machine, dramsch2020seventy, zhu2017deep, waldeland2018cnn, wu2019faultseg3d, hall2016facies, bressan2020lithology, deng2023k2, lin2024geogalactica}.
Recent foundation models extend this line of work from task-specific predictors toward language- and vision-language systems that can answer domain questions, follow instructions, and synthesize scientific context.
Geoscience-specific language models such as K2 and GeoGalactica adapt general large language models (LLMs) with geoscience corpora and instruction data \citep{deng2023k2, lin2024geogalactica}, while remote-sensing VLMs such as GeoChat show how multimodal instruction-following can be specialized to geospatial imagery \citep{kuckreja2024geochat}.
These systems suggest that geological artificial intelligence (AI) is moving from classification toward reasoning, but they do not directly test whether a model can reconstruct a latent geological history from a static diagram.

Controlled geological generators provide a way to make this question measurable.
Earlier generative-model work used neural networks to represent distributions of porous media and to define geological priors \citep{mosser2017porous, mosser2020waveform}.
More recently, conversational systems have shown how large language models can expose decades of exploration knowledge through natural-language interfaces \citep{mosser2024explorationrobot}.
Geo-Strat-RL follows the same practical pattern: the model is not asked to describe geology in the abstract, but to recover a known geological object from controlled observations and to be evaluated against quantities that can be checked.

Stratigraphic cross sections provide a compact setting for this question.
The final image is static, but the interpretation depends on temporal ordering, superposition, cross-cutting relationships, and the distinction between deposition, deformation, erosion, and intrusion.
This makes the task different from object recognition or captioning: the model must infer a process history whose steps are only indirectly visible.
General VLMs such as LLaVA-style models connect a visual encoder to an instruction-following language model \citep{liu2023llava}; recent Qwen-VL models provide strong open multimodal baselines \citep{qwen2025qwen25vl}.
Qwen3-VL further emphasizes stronger multimodal reasoning and multi-level visual feature integration \citep{qwen2025qwen3vl}.
However, broad VLM benchmarks rarely isolate whether the model has learned domain-specific temporal rules rather than surface-level description.

We therefore frame stratigraphic interpretation as an RLVR problem.
The generator provides images and exact event histories, the verifier maps model output to decomposed geological rewards, and training can optimize the model against this executable feedback without human preference labels.
This follows recent reasoning-model work in which reinforcement learning improves answer-verifiable tasks \citep{shao2024deepseekmath, deepseekai2025deepseekr1}, but we apply the idea to visual geological event reconstruction.
To make the approach practical on open VLMs, we train low-rank adaptation (LoRA) adapters \citep{hu2021lora} with Hugging Face Transformer Reinforcement Learning (TRL)'s Group Relative Policy Optimization (GRPO) implementation, which supports custom reward functions and multimodal GRPO training \citep{vonwerra2020trl, hftrlgrpo2026}.

This paper studies whether current VLMs can infer geological event histories from synthetic cross sections without relying on rendered layer names, arrows, or other visual annotations.
The objective is not to produce the most realistic geological diagrams or seismic representations.
Rather, Geo-Strat-RL uses geology as a compact, high-ambiguity reasoning domain in which generalisation across geoscience domains from RLVR can be studied under controlled supervision: the observations require temporal and structural reasoning, the answer is verifiable, and the same latent event history can be rendered through different representation domains.
The benchmark is intentionally narrow: each image is generated by a controlled simulator with a known event sequence, and each model response is scored against a compact JavaScript Object Notation (JSON) target.

\begin{figure}[!b]
  \centering
  \includegraphics[width=\columnwidth]{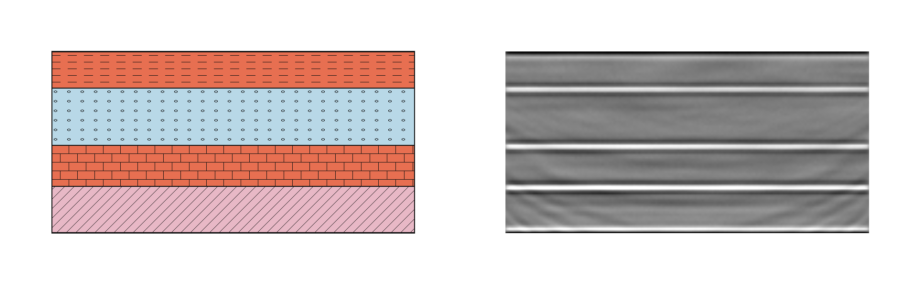}
  \caption{A simple paired held-out scene rendered as a diagram observation and as a seismic-style amplitude observation. Both panels correspond to the same hidden geological history and compact target.}
  \label{fig:seismic-transfer-pair-simple}
\end{figure}

\newpage
\subsection*{Contributions}
\begin{itemize}
  \item A controlled geological generator that produces stratigraphic observations together with compact visible-evidence event histories.
  \item An executable verifier that decomposes model performance into format, chronology, event identity, deposition, and structural-attribute components.
  \item A training and evaluation workflow that uses generated examples and verifiable rewards to improve open VLM geological reconstruction.
  \item A fixed no-label train/validation/test fixture suite with component-wise confidence intervals for diagram and seismic-style observations.
  \item Paired diagram/seismic-style experiments that test whether policies trained on one observation domain transfer to the other for the same latent geological histories.
\end{itemize}

\subsection*{Artifacts}
Repository, dataset, model, and adapter links are collected in \cref{app:benchmark-details}.

\section{Methodology}
\label{sec:methodology}

\subsection{Task Definition}
\label{subsec:task}

Given a cross-section image $x$, the model must produce a compact JSON object $y$ containing a chronological event sequence.
Each event is represented with a short event code and attributes:
\begin{align}
  y = \{\texttt{"seq"}: [e_1, e_2, \ldots, e_T]\}.
\end{align}
The same compact target can be paired with different renderings of the hidden scene; \cref{fig:seismic-transfer-pair-simple} shows a simple diagram/seismic-style pair used in the transfer setting.
The target chronology is generated directly by the simulator and then normalized to include only evidence visible in the final rendered image, with an explicit initial \texttt{basement} unit retained as a fixed convention.
The compact schema supports five event codes: deposition \texttt{D}, tilting \texttt{T}, faulting \texttt{F}, erosion or angular unconformity \texttt{E}, and intrusion \texttt{I}.
Each deposition event names \texttt{basement} or one layer; deposition events are not grouped.
Tilting records affected layers and clockwise or counterclockwise direction.
Faulting records a fault identifier and normal or reverse motion.
Erosion records an angular-unconformity surface and visible truncated layers or structures.
Intrusion records a dike identifier, preserved start and end points, and visible cut layers.
For intrusions whose endpoints fall inside the stratigraphic column, the layer endpoint is approximated from generated vertical endpoint metadata rather than inferred independently from rendered pixels; it is therefore a controlled target convention, not a claim about human image-picking precision.
The event letters are a deliberate interface choice rather than geological notation: fixed short codes reduce completion length, lower truncation risk, simplify deterministic parsing, and keep the verifier stable across model families.
Fault events intentionally omit affected-layer lists because such lists are ambiguous in final static images after erosion and later deposition; the benchmark instead scores fault chronology, identity, and normal/reverse sense.

\subsection{Synthetic Generator}
\label{subsec:generator}

The generator samples a chronological sequence and then renders the final state.
It starts from an explicit basement unit, deposits a lower package of layers, optionally tilts that package, adds complexity-dependent finite faults and intrusions, applies erosion or an angular unconformity, and deposits younger layers.
For complexity levels 4 and 5, an intrusion is guaranteed later in the program if none was sampled before erosion.
The randomized generator uses 4--9 layers and cycles through five complexity levels when building training rows.
Lithologies are drawn from a small stylized set.
Diagram colors and textures are not semantic identifiers: basement has a fixed diagnostic style, while non-basement depositional units and intrusions receive seed-deterministic randomized styles with contrast constraints between neighboring units.

The geometric model is deliberately textbook-like.
Deposition creates polygonal units with finite visible thickness.
Tilting rotates existing layers around the lower frame.
Faulting uses a finite displacement field in fault-local coordinates with along-fault decay, across-fault attenuation, and a narrow core transition; affected layers and existing intrusions are split and deformed rather than simply overprinted by a fault line.
Erosion clips existing layers and intrusions below a sloping unconformity surface, and intrusion inserts an irregular dike polygon through the current stratigraphy.
After construction, the scene is vertically scaled into the frame, deformation voids are assigned either to basement exposure or to an upward extension of the youngest visible layer, and the final input image is rendered without layer names, arrows, or event markers.
The event vocabulary follows standard sequence-stratigraphic and structural-geology concepts, but the renderer is intentionally schematic rather than a physical forward model \citep{catuneanu2006principles, allan1989faults}.

These assumptions are intentionally restrictive.
The diagram benchmark does not model depositional facies uncertainty, ambiguous outcrop exposure, or multiple valid geological interpretations.
Instead, it provides a controlled process-history task where the simulator defines a single answer key and where failures can be attributed to visual recognition, chronological ordering, or schema compliance.

\begin{figure*}[t]
  \centering
  \includegraphics[width=\textwidth]{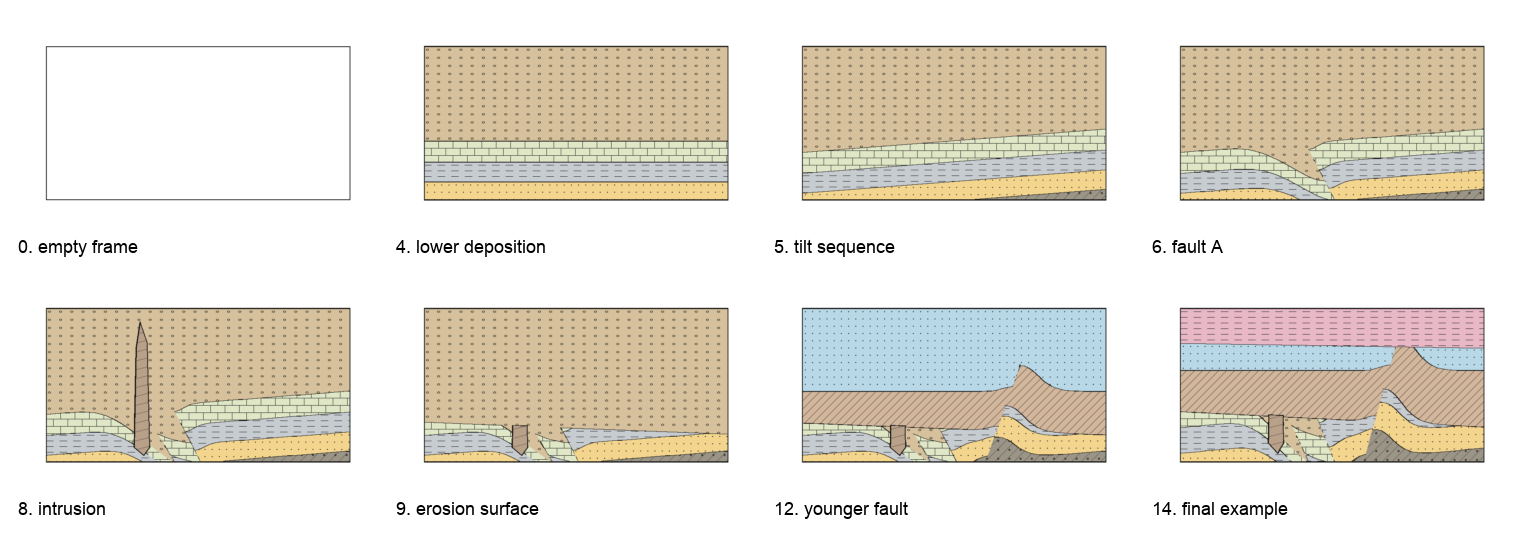}
  \caption{Construction flow for generated seed 20001. The training and evaluation images contain only the final cross section, but the simulator constructs the scene through a known chronological event sequence.}
  \label{fig:construction-flow}
\end{figure*}

\begin{figure*}[t]
  \centering
  \includegraphics[width=\textwidth]{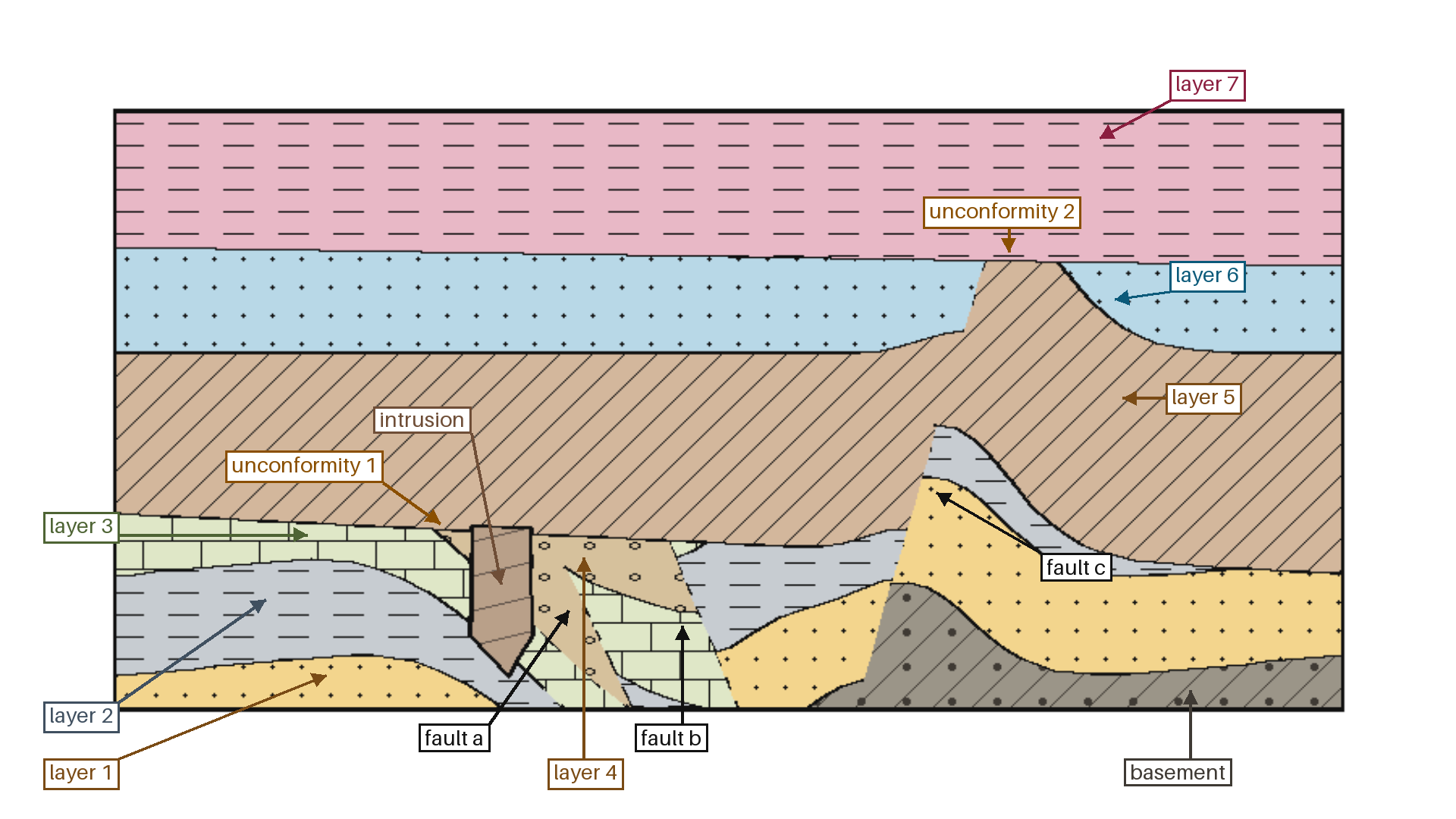}
  \caption{Enlarged annotated final cross section for generated seed 20001. This panel is for reader audit only; benchmark images do not contain labels or arrows.}
  \label{fig:annotated-final}
\end{figure*}

\noindent\textbf{Compact target example.}
For the construction flow in \cref{fig:construction-flow}, the visible-evidence target is:
\begin{jsonbox}
{
  "seq": [
    ["D", "basement"],
    ["D", "layer_1"],
    ["D", "layer_2"],
    ["D", "layer_3"],
    ["D", "layer_4"],
    ["T", ["basement", "layer_1", "layer_2", "layer_3", "layer_4"], "clockwise"],
    ["F", "fault_a", "normal"],
    ["F", "fault_b", "normal"],
    ["I", "intrusion_a", "layer_2", "angular_unconformity", ["layer_2", "layer_3", "layer_4"]],
    ["E", "angular_unconformity", ["layer_1", "layer_2", "layer_3", "layer_4", "fault_a", "fault_b", "intrusion_a"]],
    ["D", "layer_5"],
    ["D", "layer_6"],
    ["F", "fault_c", "reverse"],
    ["E", "angular_unconformity", ["layer_5", "layer_6", "fault_c"]],
    ["D", "layer_7"]
  ]
}
\end{jsonbox}

\subsection{Seismic-Style Transfer Renderer}
\label{subsec:seismic-renderer}

To test whether diagram-domain RLVR learns geological reasoning that survives a controlled observation-domain shift, we add a seismic-style renderer for the same hidden scenes.
This renderer changes only the observed data domain, not the sampled geological history or compact JSON target.
For each seed, the paired diagram and seismic-style observations therefore have the same event sequence and the same visible-evidence answer key.
The diagram observation is a colored schematic cross section, while the seismic-style observation is a synthetic acoustic amplitude section represented as an image for VLM input; \cref{fig:seismic-transfer-pair-simple} shows a minimal paired example.

The renderer first rasterizes three acoustic property fields over the final scene: P-wave velocity $v_p$, density $\rho$, and acoustic impedance $Z=\rho v_p$.
Depositional units receive deterministic acoustic impedances that increase by a fixed minimum step from one layer to the next, with only a small lithology-dependent perturbation.
This makes each depositional contact produce a visible reflection coefficient, including contacts between visually similar or locally adjacent units.
Faults, unconformities, and intrusions enter through the same final deformed geometry used by the diagram renderer; intrusions receive a high-impedance body, but their interior amplitudes are smoothed so the task cue is the body and its cut relationships rather than point-diffraction artifacts.

The velocity model is then used in a simple two-dimensional acoustic finite-difference calculation.
We place a line of surface shots and receivers over a padded model, inject a Ricker wavelet, propagate pressure waves with an absorbing boundary sponge, mute direct arrivals, and record synthetic shot gathers.
The shot gathers are mapped back to depth with a lightweight Kirchhoff-style imaging condition using source--image--receiver travel times and illumination normalization.
The final displayed seismic-style image blends this depth image with the impedance-contact image, adds controlled coherent noise and trace-gain variation, and scales amplitudes to grayscale.
This produces a depth-domain acoustic-amplitude section whose layers, faults, unconformities, and intrusions remain aligned with the paired diagram while adding basic wave-propagation and acquisition effects.
In addition to the GRPO runs performed on stratigraphic diagrams as the training domain, we additionally train one Qwen3-VL-4B adapter on the paired seismic-style observations to test reverse transfer back to stratigraphic diagrams.

\subsection{Environment and Datasets}
\label{subsec:environment}

The generator is exposed to the reinforcement learning framework TRL as a materialized Hugging Face dataset.
Each row contains a conversational prompt, a Portable Network Graphics (PNG)-encoded image, the compact answer in the \texttt{answer} field, and metadata such as seed, complexity, and number of visible events.
The same interface can generate examples on demand by advancing through deterministic seeds, so the environment is not intrinsically limited to a closed set of diagrams.
For reproducibility, and lower training-time overhead, the reported runs instead use a fixed cached training split with distinct seeded rows; this split is larger than the number of prompt rows consumed by the 2000-step training jobs.
The RLVR training runs reported here use images without rendered visual annotations, 16 generations per prompt, generation batch size 16, LoRA rank 16 with alpha 32 and dropout 0.05, target modules \texttt{q\_proj} and \texttt{v\_proj}, and 2000 GRPO training steps.
The main Qwen2.5-VL-3B and Qwen3-VL-4B adapters are trained on diagram observations; a separate Qwen3-VL-4B adapter is trained on seismic-style observations with the same GRPO recipe.
Validation and test fixtures are fixed disjoint Hugging Face datasets; exact artifact identifiers and seed ranges are listed in \cref{app:benchmark-details}.
The results in this paper report only held-out test performance.
\Cref{tab:data-composition} summarizes the fixed train, validation, and test splits.

\begin{table}[t]
\caption{Dataset composition of geological events. All reported RLVR runs use the same fixed training split. Validation and test rows report exact compact visible targets for the fixed held-out fixtures.}
\label{tab:data-composition}
\centering
\small
\resizebox{\columnwidth}{!}{%
\begin{tabular}{lrrrrrrr}
\toprule
Split & Rows & $D$ & $T$ & $F$ & $E$ & $I$ & Events/row \\
\midrule
Train & 16{,}384 & 125{,}056 & 9{,}784 & 14{,}403 & 14{,}386 & 6{,}553 & 10.39 \\
Validation & 100 & 748 & 64 & 86 & 88 & 40 & 10.26 \\
Held-out test & 100 & 747 & 59 & 88 & 88 & 40 & 10.22 \\
\bottomrule
\end{tabular}
}
\end{table}

\begin{figure*}[t]
  \centering
  \includegraphics[width=\textwidth]{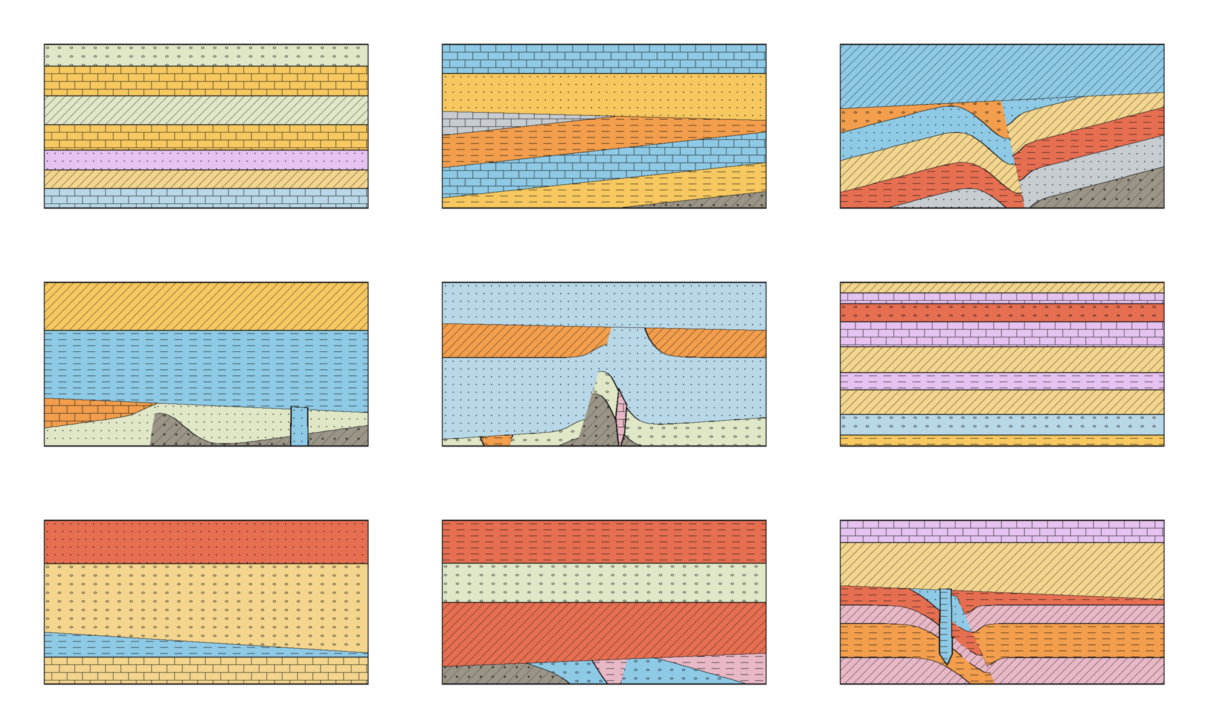}
  \caption{Examples from the synthetic dataset. The model receives only the stratigraphic diagram as an image and must recover a compact chronological event sequence.}
  \label{fig:dataset-examples}
\end{figure*}

\begin{figure*}[t]
  \centering
  \includegraphics[width=\textwidth]{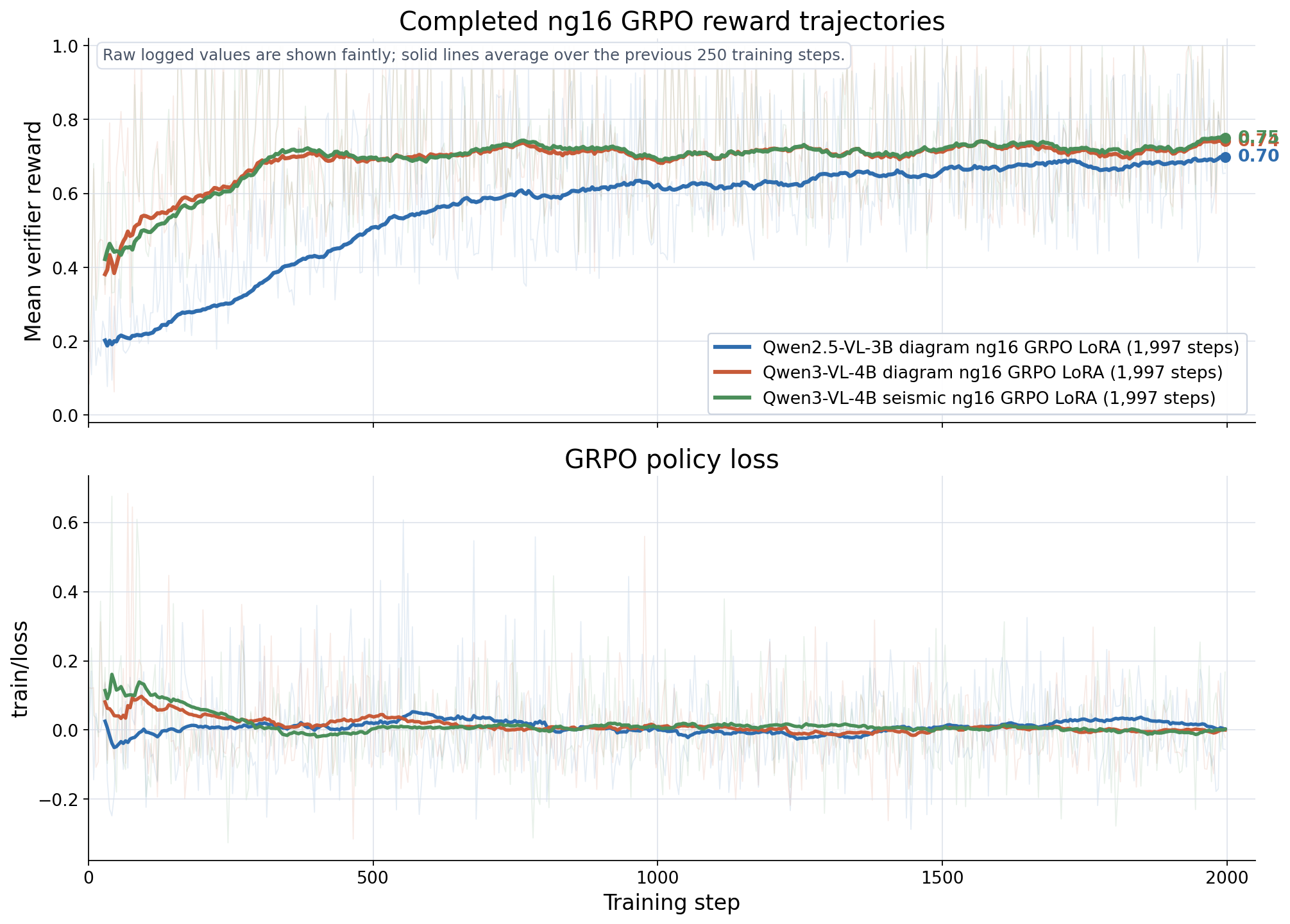}
  \caption{Training reward and loss during the completed 2{,}000-step TRL/GRPO LoRA training runs. The curves include Qwen2.5-VL-3B diagram-domain training, Qwen3-VL-4B diagram-domain training, and Qwen3-VL-4B seismic-domain training; final claims are based on held-out test rows and the transfer analysis rather than training reward alone.}
  \label{fig:grpo-reward-overlay}
\end{figure*}

\subsection{Reward Components}
\label{subsec:reward}

The prompt requires the model to emit only compact JSON with a top-level \texttt{seq} field.
The verifier first extracts and parses JSON, accepting fenced JSON only after stripping the fence.
Invalid JSON receives zero reward.
For a valid compact response, the scalar training reward $R$ is a weighted sum of ten component scores:
\begin{align}
R &= 0.10R_{\text{json}} + 0.15R_{\text{count}} + 0.25R_{\text{order}}
   + 0.10R_{\text{types}} \notag\\
  &\quad + 0.15R_{\text{dep}} + 0.10R_{\text{fault}}
   + 0.08R_{\text{intr}} + 0.04R_{\text{unc}} \notag\\
  &\quad + 0.02R_{\text{extra}} + 0.01R_{\text{prose}}.
\end{align}
Each component lies in $[0,1]$, so the weights express how much the verifier values different failure modes.
$R_{\text{json}}$ checks only that the response is parseable compact JSON.
$R_{\text{count}}$, $R_{\text{order}}$, and $R_{\text{types}}$ evaluate the coarse event history: the number of events, their chronological event-code order, and the set of event codes present.
$R_{\text{dep}}$, $R_{\text{fault}}$, $R_{\text{intr}}$, and $R_{\text{unc}}$ evaluate event-specific geological attributes for deposition, faults, intrusions, and unconformities.
The final two terms are interface and parsimony terms: $R_{\text{extra}}$ penalizes hallucinated additional events, and $R_{\text{prose}}$ penalizes explanations around the JSON object.
For an unparsable response, the verifier sets $R=0$ rather than applying these terms.
Thus a syntactically valid compact response with no geological match can still receive the format reward; this is useful during RL training because it bootstraps the required response interface, but it is not counted as geological reasoning in the geology-only analysis.
For reporting geological reconstruction independent of response-interface compliance, we define
\begin{align}
R_{\text{geo}} = \frac{1}{7}(&R_{\text{count}} + R_{\text{order}} + R_{\text{types}}
 + R_{\text{dep}} \notag\\
&+ R_{\text{fault}} + R_{\text{intr}} + R_{\text{unc}}).
\end{align}
This unweighted geology-only score is not the training objective.
It is used for cross-domain analysis because it excludes JSON validity, prose suppression, and extra-event parsimony, and therefore better reflects event-history recovery once a compact answer has been produced.
The seven terms in $R_{\text{geo}}$ are intentionally averaged with equal weight: this makes the reported geology score less sensitive to the particular RL reward weights in the training objective.
The table and log names follow these terms directly.
\texttt{json\_valid} is the parseability term, and \texttt{no\_prose} is the compact-output term.
\texttt{event\_count} penalizes missing or extra events; \texttt{event\_order} compares event codes in chronological position with an additional length penalty; and \texttt{event\_types} measures set overlap over event codes.
\texttt{deposition\_units} matches layer identifiers for \texttt{D} events.
\texttt{fault\_attributes} matches fault identifier and normal/reverse sense, while \texttt{intrusion\_attributes} and \texttt{unconformity\_attributes} match the required compact fields for intrusions and unconformities at the expected event position.
\texttt{no\_extra\_events} is one when the response has no more events than the target and otherwise decays as target length over response length.
The remaining equations define the non-attribute components.
Let $A=(a_1,\ldots,a_M)$ be the predicted compact event list and $B=(b_1,\ldots,b_N)$ be the target list.
Here $M$ and $N$ are the predicted and target lengths, $c(e)$ returns the event code of event $e$ (for example \texttt{D}, \texttt{F}, or \texttt{I}), and $\mathbf{1}[\cdot]$ is an indicator function.
The implemented component scores are:
\begin{align}
R_{\text{count}} &= \max\left(0, 1-\frac{|M-N|}{N}\right),\\
R_{\text{order}} &= \max\left(0, S_{\text{ord}} - 0.25\frac{|M-N|}{N}\right),\\
S_{\text{ord}} &= \frac{1}{N}\sum_{i=1}^{N}\mathbf{1}[i\le M \wedge c(a_i)=c(b_i)],\\
R_{\text{types}} &= \frac{|C_A\cap C_B|}{|C_B|},\\
R_{\text{extra}} &= \begin{cases}
1, & M\le N,\\
N/M, & M>N.
\end{cases}
\end{align}
The count score is one when the response has the correct length and decreases linearly as events are missing or added.
$S_{\text{ord}}$ is the fraction of target positions whose event code matches at the same chronological position; $R_{\text{order}}$ then subtracts a smaller length penalty so that a response cannot obtain a high order score simply by matching a short prefix.
For the type-overlap score, $C_A=\{c(a_i)\}_{i=1}^{M}$ and $C_B=\{c(b_i)\}_{i=1}^{N}$ are the predicted and target sets of event codes.
This rewards whether the response includes the right kinds of geological processes, independent of how many times they occur.
$R_{\text{extra}}$ is one when the response is no longer than the target and decays as $N/M$ when the model adds unsupported events.
Typed attribute scores are averaged over target events of the relevant type at their expected positions.
For each required field, scalar/string values receive 1 for exact normalized equality and 0 otherwise, while list values receive target-normalized set overlap.
The required compact fields are: \texttt{D} basement or layer id for $R_{\text{dep}}$; \texttt{F} fault id and normal/reverse sense for $R_{\text{fault}}$; \texttt{I} intrusion id, start, end, and cut-layer list for $R_{\text{intr}}$; and \texttt{E} surface type and truncated-feature list for $R_{\text{unc}}$.
If a target contains no event of a typed class, that typed attribute score is defined as 1.
For seismic-style scoring, a prediction may omit the leading basement deposition event when the target begins with basement, because basement is not always uniquely distinguishable in the amplitude-style observations.
The prose score $R_{\text{prose}}$ is 1 for a bare JSON object, 0.75 for a fenced JSON object, and 0 when extra prose surrounds the object.
The TRL run passes this verifier as a custom reward function to \texttt{GRPOTrainer}.
TRL computes rewards over generated completions and supports weighted custom reward functions; its current GRPO documentation also explicitly supports VLM training with image-text datasets \citep{hftrlgrpo2026}.
The training and evaluation scripts use the aggregate reward for optimization and log every component for analysis.
The implementation uses Hugging Face datasets for split storage \citep{lhoest2021datasets}, and PyTorch for training utilities \citep{paszke2019pytorch}.
For API-served pretrained open and closed VLM evaluations, we use OpenRouter as the primary model provider; locally trained LoRA adapters and local base checkpoints are evaluated directly with the same prompts and verifier.

\section{Results}
\label{sec:results}

\subsection{Held-Out Test Results}
\label{subsec:rl-results}

\begin{table*}[t]
  \centering
\caption{Current held-out test performance on stratigraphic diagram images from \texttt{porestar/geo-strat-rl-test}. Each metric cell reports the mean with a 95\% bootstrap confidence interval underneath. $R$ is the full verifier reward; $R_{\text{geo}}$ averages event count, event order, event types, deposition units, fault attributes, intrusion attributes, and unconformity attributes. Unless otherwise noted, evaluations use 100 examples with 5 rollouts per example; the GPT-5.5 diagram evaluation uses 3 rollouts per example because medium-reasoning API inference was substantially more expensive. Rollouts are averaged within each example before bootstrapping over held-out examples. Err. and Trunc. report rollout-level provider/error and truncation rates where current result files are included. Arrows indicate that higher scores are better; bold underlined values mark the best reported mean in each metric.}
  \label{tab:test-results}
  \scriptsize
  \resizebox{\textwidth}{!}{%
  \begin{tabular}{lllrrrrrrrrrrrrrr}
    \toprule
    Model & Class & Method & $R$ \scoreup & $R_{\text{geo}}$ \scoreup & JSON \scoreup & Count \scoreup & Order \scoreup & Types \scoreup & Dep. \scoreup & Fault \scoreup & Intr. \scoreup & Unc. \scoreup & Extra \scoreup & Prose \scoreup & Err. \scoredown & Trunc. \scoredown \\
    \midrule
    Gemma 4 31B IT & open VLM & pretrained & \scoreci{0.701}{0.664}{0.739} & \scoreci{0.616}{0.568}{0.665} & \scoreci{0.992}{0.984}{0.998} & \scoreci{0.867}{0.845}{0.888} & \scoreci{0.603}{0.554}{0.654} & \scoreci{0.836}{0.800}{0.872} & \scoreci{0.720}{0.683}{0.756} & \scoreci{0.403}{0.310}{0.499} & \scoreci{0.608}{0.513}{0.697} & \scoreci{0.275}{0.194}{0.359} & \scoreci{0.986}{0.977}{0.994} & \scoreci{0.992}{0.984}{0.998} & \bestscore{0.000} & \bestscore{0.000} \\
    Qwen3-VL-30B-A3B & open VLM & pretrained & \scoreci{0.621}{0.578}{0.662} & \scoreci{0.552}{0.503}{0.600} & \scoreci{0.872}{0.834}{0.908} & \scoreci{0.706}{0.663}{0.747} & \scoreci{0.525}{0.478}{0.573} & \scoreci{0.817}{0.777}{0.854} & \scoreci{0.665}{0.623}{0.706} & \scoreci{0.384}{0.296}{0.474} & \scoreci{0.548}{0.458}{0.637} & \scoreci{0.218}{0.142}{0.298} & \scoreci{0.768}{0.727}{0.808} & \scoreci{0.872}{0.834}{0.908} & \bestscore{0.000} & 0.020 \\
    Mistral Small 3.2 24B & open VLM & pretrained & \scoreci{0.624}{0.591}{0.656} & \scoreci{0.563}{0.523}{0.604} & \scoreci{0.966}{0.932}{0.992} & \scoreci{0.483}{0.426}{0.540} & \scoreci{0.526}{0.484}{0.568} & \scoreci{0.940}{0.905}{0.969} & \scoreci{0.740}{0.699}{0.780} & \scoreci{0.403}{0.312}{0.497} & \scoreci{0.588}{0.492}{0.682} & \scoreci{0.264}{0.181}{0.348} & \scoreci{0.659}{0.618}{0.699} & \scoreci{0.725}{0.699}{0.744} & 0.014 & 0.020 \\
    Gemini 2.5 Flash Lite & closed VLM & pretrained & \scoreci{0.691}{0.651}{0.730} & \scoreci{0.613}{0.564}{0.661} & \scoreci{0.978}{0.956}{0.996} & \scoreci{0.780}{0.742}{0.816} & \scoreci{0.620}{0.570}{0.670} & \scoreci{0.818}{0.770}{0.863} & \scoreci{0.731}{0.687}{0.773} & \scoreci{0.423}{0.329}{0.517} & \scoreci{0.615}{0.520}{0.706} & \scoreci{0.307}{0.224}{0.393} & \scoreci{0.834}{0.806}{0.862} & \scoreci{0.978}{0.956}{0.996} & \bestscore{0.000} & 0.022 \\
    GPT-5.5 \citep{openai2026gpt55system} & closed VLM & medium reasoning & \bestci{0.747}{0.705}{0.787} & \bestci{0.668}{0.615}{0.718} & \bestci{1.000}{1.000}{1.000} & \bestci{0.917}{0.898}{0.935} & \bestci{0.681}{0.626}{0.735} & \scoreci{0.867}{0.835}{0.896} & \bestci{0.756}{0.710}{0.800} & \bestci{0.435}{0.343}{0.528} & \bestci{0.627}{0.533}{0.717} & \bestci{0.391}{0.307}{0.473} & \bestci{0.991}{0.986}{0.996} & \bestci{1.000}{1.000}{1.000} & \bestscore{0.000} & \bestscore{0.000} \\
    Qwen2.5-VL-3B & open VLM & pretrained & \scoreci{0.208}{0.182}{0.234} & \scoreci{0.189}{0.163}{0.217} & \scoreci{0.404}{0.358}{0.448} & \scoreci{0.232}{0.200}{0.265} & \scoreci{0.121}{0.103}{0.140} & \scoreci{0.314}{0.276}{0.353} & \scoreci{0.139}{0.119}{0.158} & \scoreci{0.167}{0.120}{0.216} & \scoreci{0.246}{0.195}{0.300} & \scoreci{0.107}{0.068}{0.149} & \scoreci{0.332}{0.292}{0.372} & \scoreci{0.328}{0.289}{0.365} & \bestscore{0.000} & \bestscore{0.000} \\
    Qwen2.5-VL-3B & open VLM & diagram ng16 LoRA & \scoreci{0.695}{0.664}{0.724} & \scoreci{0.609}{0.568}{0.649} & \bestci{1.000}{1.000}{1.000} & \scoreci{0.812}{0.784}{0.836} & \scoreci{0.603}{0.564}{0.642} & \scoreci{0.861}{0.827}{0.892} & \scoreci{0.750}{0.719}{0.781} & \scoreci{0.408}{0.315}{0.505} & \scoreci{0.601}{0.502}{0.691} & \scoreci{0.229}{0.153}{0.309} & \scoreci{0.905}{0.885}{0.923} & \scoreci{0.750}{0.750}{0.750} & \bestscore{0.000} & \bestscore{0.000} \\
    Qwen3-VL-4B & open VLM & pretrained & \scoreci{0.311}{0.254}{0.366} & \scoreci{0.287}{0.233}{0.341} & \scoreci{0.430}{0.358}{0.502} & \scoreci{0.355}{0.295}{0.413} & \scoreci{0.256}{0.204}{0.308} & \scoreci{0.377}{0.311}{0.441} & \scoreci{0.309}{0.255}{0.366} & \scoreci{0.233}{0.172}{0.297} & \scoreci{0.290}{0.222}{0.362} & \scoreci{0.191}{0.135}{0.249} & \scoreci{0.424}{0.355}{0.496} & \scoreci{0.376}{0.308}{0.444} & \bestscore{0.000} & \bestscore{0.000} \\
    Qwen3-VL-4B & open VLM & diagram ng16 LoRA & \scoreci{0.740}{0.708}{0.772} & \scoreci{0.664}{0.622}{0.706} & \bestci{1.000}{1.000}{1.000} & \scoreci{0.906}{0.888}{0.924} & \scoreci{0.633}{0.587}{0.679} & \bestci{0.981}{0.961}{0.996} & \scoreci{0.755}{0.722}{0.788} & \scoreci{0.428}{0.336}{0.523} & \scoreci{0.618}{0.524}{0.707} & \scoreci{0.328}{0.243}{0.415} & \scoreci{0.964}{0.953}{0.973} & \bestci{1.000}{1.000}{1.000} & \bestscore{0.000} & \bestscore{0.000} \\
    \bottomrule
  \end{tabular}
  }
\end{table*}

\begin{table*}[t]
  \centering
  \caption{Held-out test performance on seismic-style observations generated from stratigraphic diagrams. Each evaluation uses 100 examples with 5 rollouts per example, averages rollouts within example, and bootstraps over example means for 95\% confidence intervals. $R$ is the full verifier reward; $R_{\text{geo}}$ averages event count, event order, event types, deposition units, fault attributes, intrusion attributes, and unconformity attributes. Err. and Trunc. report rollout-level provider/error and truncation rates. The Gemini 2.5 Flash Lite evaluation is retained for comparison but should be read with caution: even with a 16{,}384-token output cap, 82 of 500 rollouts ended with length truncation.}
  \label{tab:seismic-openrouter-results}
  \scriptsize
  \resizebox{\textwidth}{!}{%
  \begin{tabular}{lllrrrrrrrrrrrrrr}
    \toprule
    Model & Class & Method & $R$ \scoreup & $R_{\text{geo}}$ \scoreup & JSON \scoreup & Count \scoreup & Order \scoreup & Types \scoreup & Dep. \scoreup & Fault \scoreup & Intr. \scoreup & Unc. \scoreup & Extra \scoreup & Prose \scoreup & Err. \scoredown & Trunc. \scoredown \\
    \midrule
    Gemma 4 31B IT & open VLM & pretrained & \scoreci{0.627}{0.589}{0.665} & \scoreci{0.545}{0.495}{0.593} & \scoreci{0.994}{0.986}{1.000} & \scoreci{0.710}{0.680}{0.739} & \scoreci{0.483}{0.435}{0.529} & \scoreci{0.710}{0.665}{0.754} & \scoreci{0.675}{0.644}{0.705} & \scoreci{0.420}{0.327}{0.511} & \scoreci{0.603}{0.508}{0.695} & \scoreci{0.215}{0.141}{0.295} & \bestci{0.992}{0.985}{0.998} & \scoreci{0.994}{0.986}{1.000} & \bestscore{0.000} & \bestscore{0.000} \\
    Qwen3-VL-30B-A3B & open VLM & pretrained & \scoreci{0.655}{0.616}{0.693} & \scoreci{0.582}{0.537}{0.628} & \scoreci{0.928}{0.896}{0.956} & \scoreci{0.761}{0.727}{0.794} & \scoreci{0.550}{0.501}{0.598} & \scoreci{0.853}{0.815}{0.890} & \scoreci{0.679}{0.638}{0.720} & \scoreci{0.398}{0.306}{0.490} & \scoreci{0.580}{0.489}{0.668} & \scoreci{0.254}{0.180}{0.333} & \scoreci{0.868}{0.834}{0.899} & \scoreci{0.928}{0.896}{0.956} & \bestscore{0.000} & 0.022 \\
    Mistral Small 3.2 24B & open VLM & pretrained & \scoreci{0.716}{0.683}{0.747} & \scoreci{0.636}{0.592}{0.678} & \bestci{1.000}{1.000}{1.000} & \scoreci{0.823}{0.797}{0.848} & \scoreci{0.640}{0.597}{0.681} & \scoreci{0.871}{0.844}{0.897} & \bestci{0.775}{0.741}{0.807} & \scoreci{0.434}{0.342}{0.525} & \scoreci{0.600}{0.500}{0.690} & \scoreci{0.308}{0.229}{0.388} & \scoreci{0.900}{0.879}{0.921} & \scoreci{0.750}{0.750}{0.750} & \bestscore{0.000} & \bestscore{0.000} \\
    Gemini 2.5 Flash Lite & closed VLM & pretrained & \scoreci{0.560}{0.504}{0.613} & \scoreci{0.507}{0.451}{0.563} & \scoreci{0.816}{0.744}{0.882} & \scoreci{0.504}{0.436}{0.570} & \scoreci{0.476}{0.418}{0.532} & \scoreci{0.767}{0.698}{0.833} & \scoreci{0.638}{0.576}{0.698} & \scoreci{0.371}{0.286}{0.460} & \scoreci{0.539}{0.446}{0.632} & \scoreci{0.256}{0.179}{0.338} & \scoreci{0.623}{0.560}{0.685} & \scoreci{0.816}{0.744}{0.882} & \bestscore{0.000} & 0.164 \\
    GPT-5.5 \citep{openai2026gpt55system} & closed VLM & medium reasoning & \scoreci{0.656}{0.632}{0.680} & \scoreci{0.590}{0.556}{0.623} & \bestci{1.000}{1.000}{1.000} & \scoreci{0.732}{0.699}{0.764} & \scoreci{0.552}{0.520}{0.584} & \scoreci{0.773}{0.735}{0.808} & \scoreci{0.604}{0.569}{0.637} & \bestci{0.459}{0.369}{0.546} & \bestci{0.637}{0.546}{0.723} & \bestci{0.372}{0.301}{0.445} & \scoreci{0.944}{0.929}{0.959} & \bestci{1.000}{1.000}{1.000} & \bestscore{0.000} & \bestscore{0.000} \\
    Qwen2.5-VL-3B & open VLM & pretrained & \scoreci{0.192}{0.169}{0.214} & \scoreci{0.171}{0.148}{0.194} & \scoreci{0.402}{0.360}{0.444} & \scoreci{0.245}{0.215}{0.277} & \scoreci{0.090}{0.075}{0.105} & \scoreci{0.257}{0.222}{0.291} & \scoreci{0.107}{0.091}{0.123} & \scoreci{0.165}{0.120}{0.212} & \scoreci{0.244}{0.194}{0.294} & \scoreci{0.089}{0.057}{0.124} & \scoreci{0.368}{0.328}{0.408} & \scoreci{0.339}{0.303}{0.375} & \bestscore{0.000} & \bestscore{0.000} \\
    Qwen2.5-VL-3B & open VLM & diagram ng16 LoRA & \scoreci{0.655}{0.626}{0.682} & \scoreci{0.588}{0.549}{0.627} & \scoreci{0.998}{0.994}{1.000} & \scoreci{0.758}{0.735}{0.780} & \scoreci{0.492}{0.455}{0.527} & \scoreci{0.905}{0.886}{0.923} & \scoreci{0.660}{0.632}{0.688} & \scoreci{0.428}{0.336}{0.519} & \scoreci{0.600}{0.503}{0.692} & \scoreci{0.275}{0.204}{0.349} & \scoreci{0.964}{0.948}{0.978} & \scoreci{0.748}{0.745}{0.750} & \bestscore{0.000} & \bestscore{0.000} \\
    Qwen3-VL-4B & open VLM & pretrained & \scoreci{0.244}{0.199}{0.292} & \scoreci{0.220}{0.176}{0.268} & \scoreci{0.386}{0.322}{0.450} & \scoreci{0.267}{0.219}{0.316} & \scoreci{0.176}{0.135}{0.221} & \scoreci{0.336}{0.277}{0.394} & \scoreci{0.241}{0.196}{0.288} & \scoreci{0.180}{0.123}{0.239} & \scoreci{0.238}{0.176}{0.304} & \scoreci{0.105}{0.062}{0.153} & \scoreci{0.368}{0.304}{0.431} & \scoreci{0.362}{0.302}{0.422} & \bestscore{0.000} & \bestscore{0.000} \\
    Qwen3-VL-4B & open VLM & diagram ng16 LoRA & \scoreci{0.717}{0.683}{0.750} & \scoreci{0.639}{0.596}{0.682} & \bestci{1.000}{1.000}{1.000} & \scoreci{0.846}{0.822}{0.870} & \scoreci{0.610}{0.563}{0.657} & \scoreci{0.904}{0.877}{0.930} & \scoreci{0.744}{0.708}{0.778} & \scoreci{0.448}{0.356}{0.539} & \scoreci{0.605}{0.507}{0.696} & \scoreci{0.318}{0.236}{0.402} & \scoreci{0.963}{0.949}{0.977} & \bestci{1.000}{1.000}{1.000} & \bestscore{0.000} & \bestscore{0.000} \\
    Qwen3-VL-4B & open VLM & seismic ng16 LoRA & \bestci{0.745}{0.713}{0.776} & \bestci{0.670}{0.628}{0.711} & \bestci{1.000}{1.000}{1.000} & \bestci{0.863}{0.843}{0.883} & \bestci{0.663}{0.618}{0.707} & \bestci{0.974}{0.954}{0.990} & \scoreci{0.768}{0.733}{0.801} & \scoreci{0.457}{0.367}{0.547} & \scoreci{0.603}{0.506}{0.696} & \scoreci{0.360}{0.277}{0.446} & \scoreci{0.935}{0.919}{0.950} & \bestci{1.000}{1.000}{1.000} & \bestscore{0.000} & \bestscore{0.000} \\
    \bottomrule
  \end{tabular}
  }
\end{table*}

\begin{figure*}[t]
  \centering
  \includegraphics[width=\textwidth]{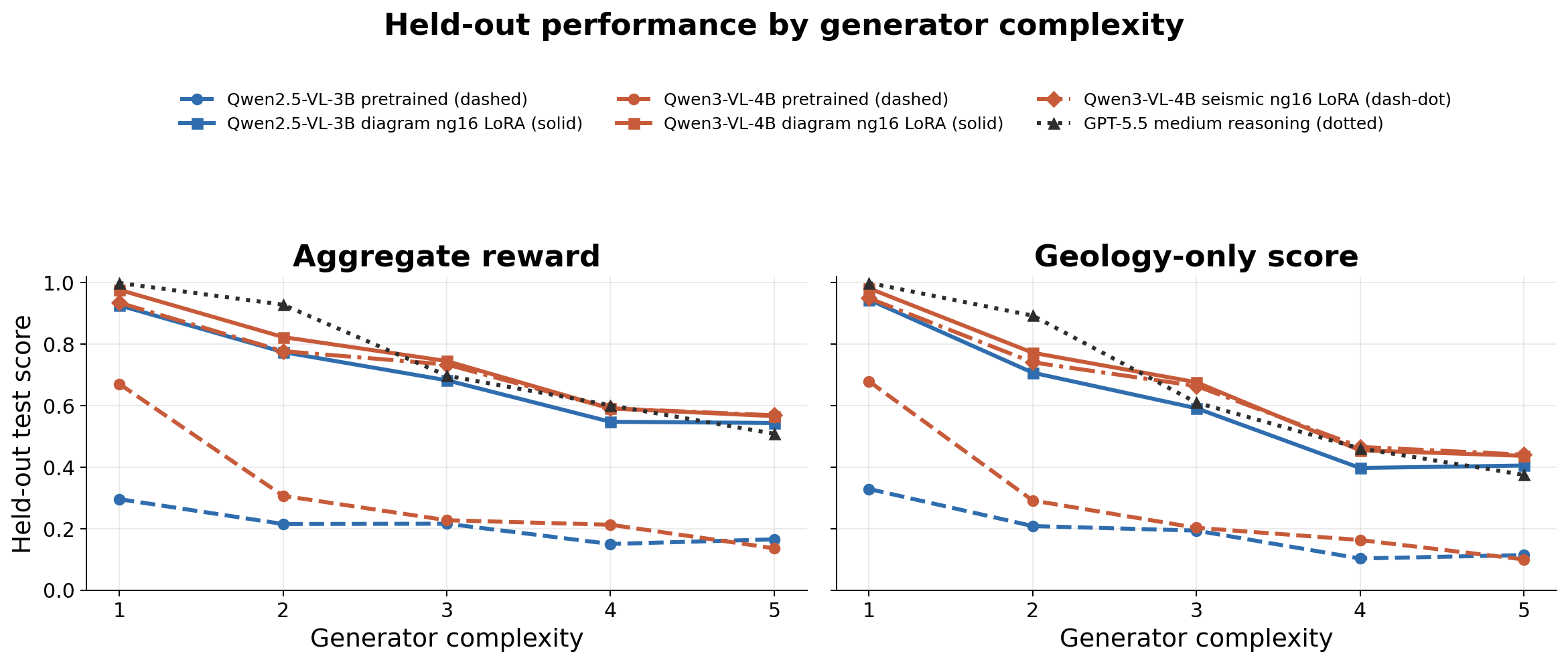}
  \caption{Held-out test performance by generator complexity on \texttt{porestar/geo-strat-rl-test}. Each complexity bucket contains 20 examples, and the same held-out examples are evaluated for the pretrained models, RLVR LoRA adapters, and OpenAI GPT-5.5 medium reasoning. Dashed lines show pretrained open models; solid lines show RLVR LoRA adapters; the dotted line shows GPT-5.5. RLVR improves both aggregate reward and geology-only score at every complexity level, while GPT-5.5 remains strongest at lower complexity and competitive at higher complexity.}
  \label{fig:complexity-breakdown}
\end{figure*}

\Cref{tab:test-results} reports held-out test performance and shows that RLVR-style optimization produces large improvements over the corresponding base models.
Each table row is one model evaluation: a specified open or closed VLM, base checkpoint, or LoRA adapter evaluated on a fixed observation domain with the same prompts, held-out examples, rollout protocol, and verifier.
Unless otherwise noted, each evaluation uses the same 100 held-out examples with 5 rollouts per example, and confidence intervals are computed by averaging rollouts within each example, then bootstrapping over the 100 example means.
For cost reasons, the GPT-5.5 diagram evaluation uses 3 rollouts per example; we disclose this exception because medium-reasoning GPT-5.5 inference was substantially more expensive than the other API evaluations.
\Cref{tab:seismic-openrouter-results} reports the corresponding seismic-style evaluations, including pretrained API-served baselines, local pretrained checkpoints, and local LoRA adapters.
For Qwen2.5-VL-3B, held-out test reward increases from 0.208 to 0.695 after diagram-domain GRPO LoRA tuning with 16 generations per prompt.
For Qwen3-VL-4B, held-out test reward increases from 0.311 to 0.740, the highest mean among the reported local open tuned evaluations.
The closed GPT-5.5 medium-reasoning diagram evaluation reaches 0.747 average reward under the same verifier after enabling JSON-object mode and omitting an output-token cap.
It is 0.007 average reward above the Qwen3-VL-4B diagram ng16 LoRA evaluation, while producing the strongest event-count, event-order, deposition-unit, fault-attribute, intrusion-attribute, unconformity-attribute, and no-extra-event scores in \cref{tab:test-results}.
For the Mistral Small 3.2 24B diagram evaluation, average reward is 0.624 with a 95\% bootstrap confidence interval of [0.591, 0.656]; this evaluation had 7 provider errors out of 500 rollouts and 10 truncated outputs out of 500 rollouts.
Mistral Small 3.2 24B has the strongest average reward among the pretrained seismic evaluations, with 0.716 [0.683, 0.747] and no provider errors or truncations in the final result file.
The strongest seismic-style evaluation overall is the Qwen3-VL-4B seismic ng16 LoRA adapter, with $R=0.745$ [0.713, 0.776] and $R_{\text{geo}}=0.670$ [0.628, 0.711].
Gemini 2.5 Flash Lite remains affected by long, non-compact seismic completions: after reevaluating with a 16{,}384-token output cap, 82 of 500 rollouts still terminate by length.
We keep this evaluation in \cref{tab:seismic-openrouter-results} because it documents observed service behavior under the same compact-JSON task, but we interpret it as a truncation-limited baseline rather than a clean estimate of model capability.
GPT-5.5 also completes cleanly on seismic-style observations, reaching 0.656 [0.632, 0.680] average reward with zero final truncations.
Both adapters reach perfect JSON validity on the diagram test split; Qwen3-VL-4B also reaches perfect no-prose compliance, while Qwen2.5-VL-3B still emits prose in some rollouts.
The remaining errors concentrate in structural attributes: fault and intrusion attributes improve moderately, while unconformity attributes remain the weakest component.
\Cref{fig:complexity-breakdown} shows that the strongest gains are not limited to the easiest examples: both adapters improve over their pretrained bases at every complexity level.
The degradation with increasing complexity remains substantial, however, which supports reporting complexity-stratified scores rather than one aggregate.

\subsection{Diagram/Seismic Domain Transfer}
\label{subsec:seismic-transfer-results}

For the transfer experiment, the most relevant question is not whether a model obeys the JSON interface but whether it extracts the geological history from a different data domain.
We therefore report a geology-only score that averages the seven content components: event count, event order, event types, deposition units, fault attributes, intrusion attributes, and unconformity attributes.
This deliberately excludes \texttt{json\_valid}, \texttt{no\_prose}, and \texttt{no\_extra\_events}, because those components mainly measure response-interface compliance and output parsimony.
The aggregate reward remains useful for training, but it mixes schema learning with geological interpretation and is less clean for a cross-modal transfer investigation.
Additional paired examples are provided in \cref{app:extended-transfer-examples}.

\begin{table*}[t]
  \centering
  \caption{Cross-domain transfer on paired held-out examples. Arrows indicate that higher values are better; bold underlined values mark the best value in each numeric column. The geology-only score averages event count, event order, event types, deposition units, fault attributes, intrusion attributes, and unconformity attributes; format and prose components are excluded. N/A indicates that no Qwen2.5-VL-3B seismic-domain adapter is reported.}
  \label{tab:seismic-transfer-geology}
  \scriptsize
  \resizebox{\textwidth}{!}{%
  \begin{tabular}{llrrrrr}
    \toprule
    Model & Test domain & Pretrained $R_{\text{geo}}$ \scoreup & Diagram LoRA $R_{\text{geo}}$ \scoreup & $\Delta R_{\text{geo}}$ \scoreup & Seismic LoRA $R_{\text{geo}}$ \scoreup & $\Delta R_{\text{geo}}$ \scoreup \\
    \midrule
    Qwen2.5-VL-3B & diagram & 0.189 & 0.609 & +0.420 & N/A & N/A \\
    Qwen2.5-VL-3B & seismic & 0.171 & 0.588 & +0.417 & N/A & N/A \\
    Qwen3-VL-4B & diagram & \bestscore{0.287} & \bestscore{0.664} & +0.377 & 0.652 & +0.365 \\
    Qwen3-VL-4B & seismic & 0.220 & 0.639 & +0.419 & \bestscore{0.670} & \bestscore{+0.449} \\
    \bottomrule
  \end{tabular}
  }
\end{table*}

\begin{figure*}[t]
  \centering
  \includegraphics[width=\textwidth]{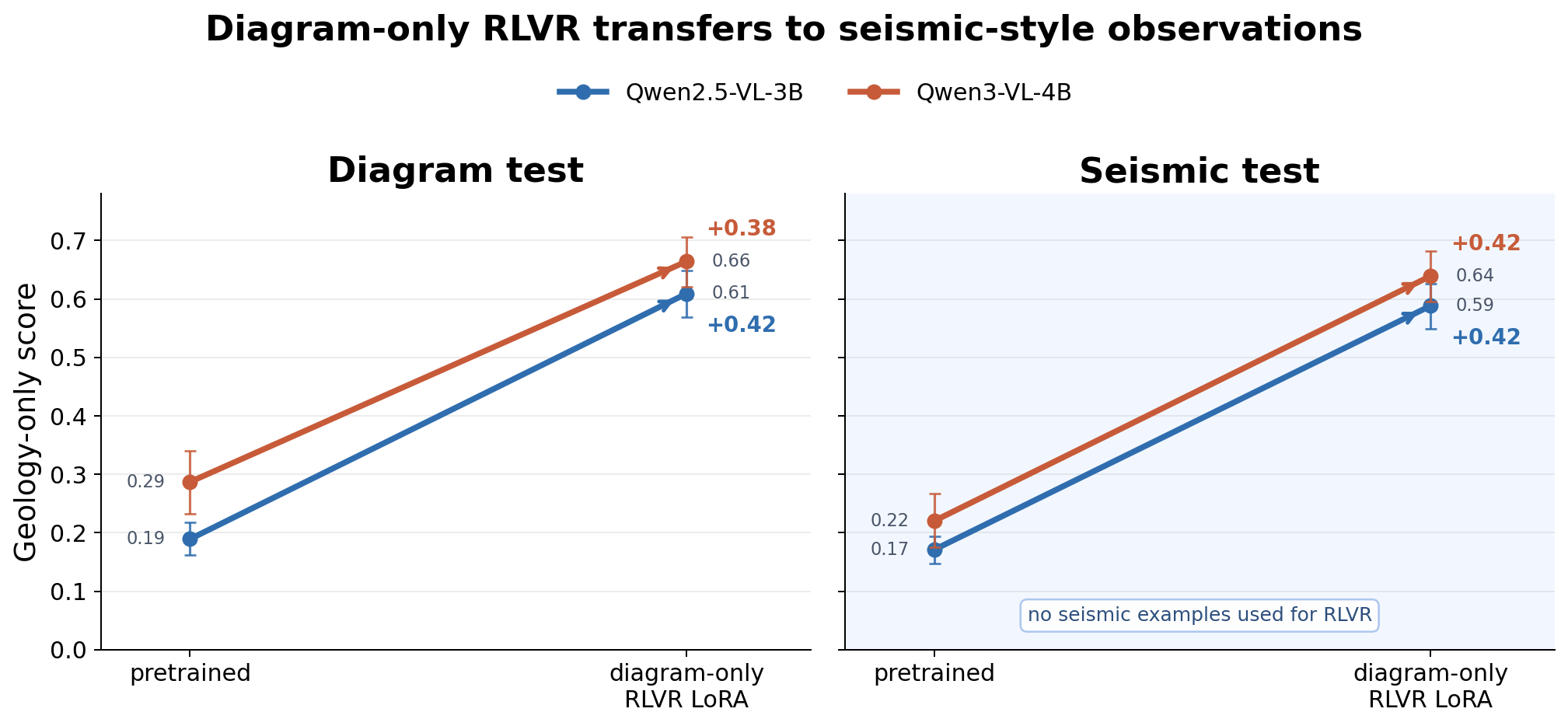}
  \caption{Before/after geology-only scores for diagram-only RLVR. Vertical bars show 95\% bootstrap confidence intervals over held-out examples. The right panel isolates transfer to seismic-style observations: no seismic-style examples are used for these two RLVR runs, yet both tuned adapters improve on the paired amplitude test.}
  \label{fig:seismic-geology-transfer}
\end{figure*}

\begin{table*}[t]
  \centering
  \caption{Detailed geology-component scores on seismic-style held-out examples. Arrows indicate that higher scores are better; bold underlined values mark the best non-delta method score in each component column. Diagram-domain LoRA adapters improve every geological component on the seismic-style test, and the Qwen3-VL-4B seismic-domain adapter further improves seismic event count, event order, event types, deposition units, fault attributes, and unconformity attributes.}
  \label{tab:seismic-transfer-components}
  \scriptsize
  \resizebox{\textwidth}{!}{%
  \begin{tabular}{llrrrrrrr}
    \toprule
    Model & Method & Count \scoreup & Order \scoreup & Types \scoreup & Dep. \scoreup & Fault \scoreup & Intr. \scoreup & Unc. \scoreup \\
    \midrule
    Qwen2.5-VL-3B & pretrained & 0.245 & 0.090 & 0.257 & 0.107 & 0.165 & 0.244 & 0.089 \\
    Qwen2.5-VL-3B & diagram LoRA & 0.758 & 0.492 & 0.905 & 0.660 & 0.428 & 0.600 & 0.275 \\
    Qwen2.5-VL-3B & delta & +0.513 & +0.402 & +0.649 & +0.553 & +0.263 & +0.356 & +0.186 \\
    \midrule
    Qwen3-VL-4B & pretrained & 0.267 & 0.176 & 0.336 & 0.241 & 0.180 & 0.238 & 0.105 \\
    Qwen3-VL-4B & diagram LoRA & 0.846 & 0.610 & 0.904 & 0.744 & 0.448 & \bestscore{0.605} & 0.318 \\
    Qwen3-VL-4B & diagram delta & +0.580 & +0.434 & +0.568 & +0.503 & +0.268 & +0.367 & +0.213 \\
    Qwen3-VL-4B & seismic LoRA & \bestscore{0.863} & \bestscore{0.663} & \bestscore{0.974} & \bestscore{0.768} & \bestscore{0.457} & 0.603 & \bestscore{0.360} \\
    Qwen3-VL-4B & seismic delta & +0.596 & +0.487 & +0.638 & +0.527 & +0.277 & +0.365 & +0.255 \\
    \bottomrule
  \end{tabular}
  }
\end{table*}

\Cref{tab:seismic-transfer-geology,fig:seismic-geology-transfer} show that diagram-domain RLVR improves geology-only scores on both the original diagram domain and the synthetic seismic-style observation domain.
\Cref{tab:seismic-transfer-geology} also includes the reverse experiment: a Qwen3-VL-4B adapter trained on seismic-style observations reaches the strongest seismic geology-only score, 0.670, and transfers back to diagram observations at 0.652.
\Cref{tab:seismic-transfer-components} shows that this is not just schema transfer; on seismic-style observations, Qwen3 diagram-domain training improves event count by 0.580, event types by 0.568, deposition units by 0.503, and intrusion attributes by 0.367, while seismic-domain training gives the strongest event order and unconformity scores.
The OpenAI GPT-5.5 medium-reasoning evaluations in \cref{tab:test-results,tab:seismic-openrouter-results} reach 0.747 average reward on diagrams and 0.656 average reward on seismic-style observations.
All 500 GPT-5.5 seismic rollouts finish with valid compact JSON and application programming interface (API) finish reason \texttt{stop}.
Qwen2.5-VL-3B improves from 0.171 to 0.588 geology-only score after diagram-domain RLVR.
The conservative conclusion is therefore that the LoRA adapters provide evidence for controlled cross-domain geological transfer in both directions, with the domain-matched Qwen3 seismic adapter ending highest on seismic and retaining most of the gain when evaluated on diagrams.

\section{Discussion}
\label{sec:discussion}

The results support four observations.
First, strong pretrained VLMs can recover some geological structure from diagrams without visual annotations, but performance varies substantially by model class and response format.
Gemma 4 31B IT is the best open model we evaluated by average reward and reaches near-perfect JSON and no-prose compliance on the 5-rollout evaluation.
The completed OpenAI GPT-5.5 medium-reasoning run is the strongest closed-model comparison currently reported here; however, its remaining errors still concentrate in the same structural attributes, especially faults and unconformities on the highest-complexity scenes.
This strong closed-model result should not be interpreted as evidence that small task-specific RLVR is already the main driver of frontier-model performance on this benchmark.
Recent reasoning-model reports show that large-scale reinforcement learning on verifiable tasks can improve general reasoning, while multimodal systems such as Kimi k1.5 and Kimi K2.5 combine RL with multimodal or joint text-vision training \citep{deepseekai2025deepseekr1, kimi2025k15, kimi2026k25}.
At the same time, recent VLM reports attribute gains to stronger visual processing and multimodal reasoning machinery, including dynamic-resolution vision transformers in Qwen2.5-VL and multi-level vision transformer (ViT) feature fusion in Qwen3-VL \citep{qwen2025qwen25vl, qwen2025qwen3vl}.
The GPT-5.5 result is therefore consistent with at least two nonexclusive explanations: broad frontier-scale RL/post-training over many verifiable tasks, and substantially stronger visual recognition and visual-temporal reasoning without any geology-specific adaptation.
It is not known which RLVR environments are used in the GPT-5.5 training recipe, and some of the improvement could come from geology-relevant tasks or specific geological training environments.
The present experiments cannot separate these mechanisms, and require deeper investigation of training recipes and ablations to determine how much of the improvement comes from general training versus geology-specific RLVR, as well as how much of the improvement comes from visual recognition versus temporal reasoning.

Second, the GRPO LoRA runs substantially improve smaller open VLMs.
Among the local open base/LoRA rows, the Qwen3-VL-4B diagram LoRA adapter has the highest average reward and strongest event-count, event-order, event-type, deposition, fault, intrusion, unconformity, no-extra-event, and no-prose means.
This suggests that executable geological rewards can shape both the response interface and parts of the visual-temporal interpretation.
However, the perfect format scores also show that some reward improvement comes from learning the compact schema, so future ablations should separate schema learning from event-history reasoning.

Third, structural attributes remain the hardest part of the task.
Even the best diagram-domain GRPO LoRA row reaches only 0.328 on unconformity attributes and 0.428 on fault attributes.
This is consistent with the qualitative difficulty of deciding which structures are visibly truncated, which layers a dike cuts, and which units are affected by younger faults after erosion and redeposition.
Future benchmark passes should keep the same prompt, reward, JSON-mode, and no-output-cap protocol so that closed-model comparisons remain complete and auditable.
This also means that there is significant room for improvement in RLVR training recipes, model architectures, and pretraining data that can improve structural reasoning and interpretation of geological relationships.
The availability of the benchmark environment will allow future updates to the test dataset to prevent overfitting and to track progress on this challenging aspect of geological reasoning.

Fourth, the seismic-style transfer experiments are the first evidence that the learned behavior is not strongly tied to the stratigraphic diagram domain.
The results of applying the diagram-pretrained models to seismic-style observations show that the same geological event structure can be extracted from a different visual domain, even without any seismic-specific training examples.
This does not prove robust seismic interpretation: the renderer deliberately makes contacts and structural relationships distinguishable, and it is still generated from the same symbolic scene model.
The result should therefore be read as evidence about reuse of verifier-trained reasoning concepts across controlled observation formats, not as a claim that the model has learned an operational seismic-interpretation workflow.
It supports the narrower hypothesis that verifier-driven training can teach reusable geological event structure across controlled observation domains, not just a response template.
This raises the question of which other verifiable domains and observation formats can share geological reasoning concepts, and how far such transfer can be pushed toward real-world geological data.
The ultimate goal is not to reproduce synthetic benchmarks, but to learn geological reasoning useful for real geoscience tasks where verification is difficult.

\subsection*{Limitations}
This benchmark is synthetic, stylized, and intentionally restricted to a small set of event types.
The generated diagrams and seismic-style observations do not capture the full ambiguity, acquisition physics, processing artifacts, noise distributions, or interpretive richness of field geology and real seismic interpretation.
Although the seismic-style renderer includes acoustic finite-difference propagation, surface acquisition, absorbing padding, simple depth imaging, coherent noise, and trace-gain variation, it should be treated as a controlled transfer proxy rather than as a realistic seismic processing workflow.
The setup also differs from how geoscientists normally interpret outcrops or seismic data.
Sequence-stratigraphic and subsurface interpretation commonly integrate multiple evidence sources, including outcrop analogs, cores, well logs, regional geological context, and multiple seismic sections or volumes with different imaging quality \citep{catuneanu2009standardization, posamentier2003seismicgeomorphology}.
Real interpretations are often revised through this integrated comparison process, whereas Geo-Strat-RL gives the model a single isolated image and evaluates it against one simulator-defined visible-evidence target.
This design is useful for controlled RLVR experiments but should not be confused with the broader expert workflow used for field, reservoir, or exploration interpretation.

\subsection*{Future Work}
Future work should add human geological baselines, richer depositional environments, uncertainty-aware targets, additional VLM families, and ablations that separate visual recognition errors from temporal-ordering errors.
The benchmark also makes a natural test bed for interpretability.
Because every response is decomposed into event-count, order, type, deposition, fault, intrusion, and unconformity scores, failures can be tied to specific geological operations rather than only to an aggregate reward.
Future work could combine component-level errors with VLM interpretability methods, such as visual-token attribution, attention or activation analysis, patch ablations, and representation probes, to test which image regions and visual relationships drive each predicted event.
When models expose visible rationales or chain-of-thought traces under an appropriate evaluation protocol, those traces could also be inspected against the executable event history to distinguish perceptual failures from failures of chronological or cross-cutting reasoning.

\section*{Impact Statement}

This paper studies synthetic evaluation and training environments for geological reasoning with vision-language models.
The primary intended impact is improved measurement of domain-specific visual reasoning, with potential downstream relevance to geoscience education and model-evaluation methodology.
The benchmark is not intended for operational subsurface interpretation or safety-critical geological decision making.

\section*{Acknowledgments}

I thank Håkon Nese and Ashley Clarke for their reviews and for providing improvement suggestions for the manuscript and method.

\bibliographystyle{icml2026}
\bibliography{references}

@article{kimi2026k25,
  author = {{Kimi Team}},
  title = {Kimi K2.5: Visual Agentic Intelligence},
  journal = {arXiv preprint arXiv:2602.02276},
  year = {2026},
  url = {https://arxiv.org/abs/2602.02276}
}

@misc{openai2026gpt55system,
  author = {{OpenAI}},
  title = {{GPT-5.5} System Card},
  year = {2026},
  url = {https://openai.com/index/gpt-5-5-system-card/}
}

@article{kimi2025k15,
  author = {{Kimi Team}},
  title = {Kimi k1.5: Scaling Reinforcement Learning with {LLMs}},
  journal = {arXiv preprint arXiv:2501.12599},
  year = {2025},
  url = {https://arxiv.org/abs/2501.12599}
}

@article{deng2023k2,
  author = {Deng, Cheng and Zhang, Tianhang and He, Zhongmou and Xu, Yi and Chen, Qiyuan and Shi, Yuanyuan and Fu, Luoyi and Zhang, Weinan and Wang, Xinbing and Zhou, Chenghu and Lin, Zhouhan and He, Junxian},
  title = {{K2}: A Foundation Language Model for Geoscience Knowledge Understanding and Utilization},
  journal = {arXiv preprint arXiv:2306.05064},
  year = {2023},
  url = {https://arxiv.org/abs/2306.05064}
}

@article{lin2024geogalactica,
  author = {Lin, Zhouhan and Deng, Cheng and Zhou, Le and Zhang, Tianhang and Xu, Yi and Xu, Yutong and He, Zhongmou and Shi, Yuanyuan and Dai, Beiya and Song, Yunchong and Zeng, Boyi and Chen, Qiyuan and Miao, Yuxun and Xue, Bo and Wang, Shu and Fu, Luoyi and Zhang, Weinan and He, Junxian and Zhu, Yunqiang and Wang, Xinbing and Zhou, Chenghu},
  title = {{GeoGalactica}: A Scientific Large Language Model in Geoscience},
  journal = {arXiv preprint arXiv:2401.00434},
  year = {2024},
  url = {https://arxiv.org/abs/2401.00434}
}

@inproceedings{kuckreja2024geochat,
  author = {Kuckreja, Kartik and Danish, Muhammad Shahzad and Naseer, Muzammal and Das, Abhijit and Khan, Salman and Khan, Fahad Shahbaz},
  title = {{GeoChat}: Grounded Large Vision-Language Model for Remote Sensing},
  booktitle = {Proceedings of the IEEE/CVF Conference on Computer Vision and Pattern Recognition},
  year = {2024},
  url = {https://openaccess.thecvf.com/content/CVPR2024/html/Kuckreja_GeoChat_Grounded_Large_Vision-Language_Model_for_Remote_Sensing_CVPR_2024_paper.html}
}

@inproceedings{liu2023llava,
  author = {Liu, Haotian and Li, Chunyuan and Wu, Qingyang and Lee, Yong Jae},
  title = {Visual Instruction Tuning},
  booktitle = {Advances in Neural Information Processing Systems},
  volume = {36},
  year = {2023},
  url = {https://arxiv.org/abs/2304.08485}
}

@article{hu2021lora,
  author = {Hu, Edward J. and Shen, Yelong and Wallis, Phillip and Allen-Zhu, Zeyuan and Li, Yuanzhi and Wang, Shean and Wang, Lu and Chen, Weizhu},
  title = {{LoRA}: Low-Rank Adaptation of Large Language Models},
  journal = {arXiv preprint arXiv:2106.09685},
  year = {2021},
  url = {https://arxiv.org/abs/2106.09685}
}

@article{shao2024deepseekmath,
  author = {Shao, Zhihong and Wang, Peiyi and Zhu, Qihao and Xu, Runxin and Song, Junxiao and Bi, Xiao and Zhang, Haowei and Zhang, Mingchuan and Li, Y. K. and Wu, Y. and Guo, Daya},
  title = {{DeepSeekMath}: Pushing the Limits of Mathematical Reasoning in Open Language Models},
  journal = {arXiv preprint arXiv:2402.03300},
  year = {2024},
  url = {https://arxiv.org/abs/2402.03300}
}

@article{deepseekai2025deepseekr1,
  author = {{DeepSeek-AI}},
  title = {{DeepSeek-R1}: Incentivizing Reasoning Capability in {LLMs} via Reinforcement Learning},
  journal = {arXiv preprint arXiv:2501.12948},
  year = {2025},
  url = {https://arxiv.org/abs/2501.12948}
}

@article{qwen2025qwen25vl,
  author = {{Qwen Team}},
  title = {{Qwen2.5-VL} Technical Report},
  journal = {arXiv preprint arXiv:2502.13923},
  year = {2025},
  url = {https://arxiv.org/abs/2502.13923}
}

@article{qwen2025qwen3vl,
  author = {{Qwen Team}},
  title = {{Qwen3-VL} Technical Report},
  journal = {arXiv preprint arXiv:2511.21631},
  year = {2025},
  url = {https://arxiv.org/abs/2511.21631}
}

@misc{vonwerra2020trl,
  author = {von Werra, Leandro and Belkada, Younes and Tunstall, Lewis and Beeching, Edward and Thrush, Tristan and Lambert, Nathan and Huang, Shengyi and Rasul, Kashif and Gallou{\'e}dec, Quentin},
  title = {{TRL}: Transformers Reinforcement Learning},
  year = {2020},
  url = {https://github.com/huggingface/trl}
}

@misc{hftrlgrpo2026,
  author = {{Hugging Face}},
  title = {{TRL} {GRPOTrainer} Documentation},
  year = {2026},
  url = {https://huggingface.co/docs/trl/en/grpo_trainer}
}

@inproceedings{lhoest2021datasets,
  author = {Lhoest, Quentin and Villanova del Moral, Albert and Jernite, Yacine and Thakur, Abhishek and von Platen, Patrick and Patil, Suraj and Chaumond, Julien and Drame, Mariama and Plu, Julien and Tunstall, Lewis and Davison, Joe and Sajjad, Hassan and Chhablani, Gunjan and Malik, Bhavitvya and Brandeis, Simon and Le Scao, Teven and Sanh, Victor and Xu, Canwen and Patry, Nicolas and McMillan-Major, Angelina and Schmid, Philipp and Gugger, Sylvain and Delangue, Clement and Matussiere, Theo and Debut, Lysandre and Bekman, Stas and Cistac, Pierric and Goehringer, Thibault and Mustar, Victor and Lagunas, Francois and Rush, Alexander M. and Wolf, Thomas},
  title = {Datasets: A Community Library for Natural Language Processing},
  booktitle = {Proceedings of the 2021 Conference on Empirical Methods in Natural Language Processing: System Demonstrations},
  pages = {175--184},
  year = {2021},
  url = {https://aclanthology.org/2021.emnlp-demo.21/}
}

@inproceedings{paszke2019pytorch,
  author = {Paszke, Adam and Gross, Sam and Massa, Francisco and Lerer, Adam and Bradbury, James and Chanan, Gregory and Killeen, Trevor and Lin, Zeming and Gimelshein, Natalia and Antiga, Luca and Desmaison, Alban and Kopf, Andreas and Yang, Edward and DeVito, Zachary and Raison, Martin and Tejani, Alykhan and Chilamkurthy, Sasank and Steiner, Benoit and Fang, Lu and Bai, Junjie and Chintala, Soumith},
  title = {{PyTorch}: An Imperative Style, High-Performance Deep Learning Library},
  booktitle = {Advances in Neural Information Processing Systems},
  volume = {32},
  year = {2019},
  url = {https://papers.neurips.cc/paper_files/paper/2019/hash/bdbca288fee7f92f2bfa9f7012727740-Abstract.html}
}

@book{catuneanu2006principles,
  author = {Catuneanu, Octavian},
  title = {Principles of Sequence Stratigraphy},
  publisher = {Elsevier},
  year = {2006}
}

@article{catuneanu2009standardization,
  author = {Catuneanu, Octavian and Abreu, Victor and Bhattacharya, Janok P. and Blum, Michael D. and Dalrymple, Robert W. and Eriksson, Patrick G. and Fielding, Christopher R. and Fisher, William L. and Galloway, William E. and Gibling, Martin R. and Giles, Katherine A. and Holbrook, John M. and Jordan, Ronald and Kendall, Christopher G. St. C. and Macurda, Bruce and Martinsen, Ole J. and Miall, Andrew D. and Neal, Jack E. and Nummedal, Dag and Pomar, Luis and Posamentier, Henry W. and Pratt, Brian R. and Sarg, J. Frederick and Shanley, Keith W. and Steel, Ronald J. and Strasser, Andr{\'e} and Tucker, Maurice E. and Winker, Charles},
  title = {Towards the Standardization of Sequence Stratigraphy},
  journal = {Earth-Science Reviews},
  year = {2009},
  volume = {92},
  number = {1--2},
  pages = {1--33},
  doi = {10.1016/j.earscirev.2008.10.003}
}

@article{posamentier2003seismicgeomorphology,
  author = {Posamentier, Henry W. and Kolla, Venkatarathnan},
  title = {Seismic Geomorphology and Stratigraphy of Depositional Elements in Deep-Water Settings},
  journal = {Journal of Sedimentary Research},
  year = {2003},
  volume = {73},
  number = {3},
  pages = {367--388},
  doi = {10.1306/111302730367}
}

@article{allan1989faults,
  author = {Allan, U. S.},
  title = {Model for Hydrocarbon Migration and Entrapment within Faulted Structures},
  journal = {AAPG Bulletin},
  year = {1989},
  volume = {73},
  number = {7},
  pages = {803--811}
}

@article{bergen2019machine,
  author = {Bergen, Karianne J. and Johnson, Paul A. and de Hoop, Maarten V. and Beroza, Gregory C.},
  title = {Machine Learning for Data-Driven Discovery in Solid Earth Geoscience},
  journal = {Science},
  year = {2019},
  volume = {363},
  number = {6433},
  pages = {eaau0323},
  doi = {10.1126/science.aau0323}
}

@article{dramsch2020seventy,
  author = {Dramsch, Jesper S{\"o}ren},
  title = {70 Years of Machine Learning in Geoscience in Review},
  journal = {Advances in Geophysics},
  year = {2020},
  volume = {61},
  pages = {1--55},
  doi = {10.1016/bs.agph.2020.08.002}
}

@article{zhu2017deep,
  author = {Zhu, Xiao Xiang and Tuia, Devis and Mou, Lichao and Xia, Gui-Song and Zhang, Liangpei and Xu, Feng and Fraundorfer, Friedrich},
  title = {Deep Learning in Remote Sensing: A Comprehensive Review and List of Resources},
  journal = {IEEE Geoscience and Remote Sensing Magazine},
  year = {2017},
  volume = {5},
  number = {4},
  pages = {8--36},
  doi = {10.1109/MGRS.2017.2762307}
}

@article{waldeland2018cnn,
  author = {Waldeland, Anders U. and Jensen, Are Charles and Gelius, Leiv-J. and Solberg, Anne H. Schistad},
  title = {Convolutional Neural Networks for Automated Seismic Interpretation},
  journal = {The Leading Edge},
  year = {2018},
  volume = {37},
  number = {7},
  pages = {529--537},
  doi = {10.1190/tle37070529.1}
}

@article{wu2019faultseg3d,
  author = {Wu, Xinming and Liang, Luming and Shi, Yunzhi and Fomel, Sergey},
  title = {{FaultSeg3D}: Using Synthetic Data Sets to Train an End-to-End Convolutional Neural Network for 3D Seismic Fault Segmentation},
  journal = {Geophysics},
  year = {2019},
  volume = {84},
  number = {3},
  pages = {IM35--IM45},
  doi = {10.1190/geo2018-0646.1}
}

@article{hall2016facies,
  author = {Hall, Brendon},
  title = {Facies Classification Using Machine Learning},
  journal = {The Leading Edge},
  year = {2016},
  volume = {35},
  number = {10},
  pages = {906--909},
  doi = {10.1190/tle35100906.1}
}

@article{bressan2020lithology,
  author = {Bressan, Thiago Santi and de Souza, Marcelo Kehl and Girelli, Tiago J. and Chemale, Farid},
  title = {Evaluation of Machine Learning Methods for Lithology Classification Using Geophysical Data},
  journal = {Computers \& Geosciences},
  year = {2020},
  volume = {139},
  pages = {104475},
  doi = {10.1016/j.cageo.2020.104475}
}

@article{mosser2017porous,
  author = {Mosser, Lukas and Dubrule, Olivier and Blunt, Martin J.},
  title = {Reconstruction of Three-Dimensional Porous Media Using Generative Adversarial Neural Networks},
  journal = {Physical Review E},
  year = {2017},
  volume = {96},
  number = {4},
  pages = {043309},
  doi = {10.1103/PhysRevE.96.043309},
  url = {https://arxiv.org/abs/1704.03225}
}

@article{mosser2020waveform,
  author = {Mosser, Lukas and Dubrule, Olivier and Blunt, Martin J.},
  title = {Stochastic Seismic Waveform Inversion Using Generative Adversarial Networks as a Geological Prior},
  journal = {Mathematical Geosciences},
  year = {2020},
  volume = {52},
  number = {1},
  pages = {53--79},
  doi = {10.1007/s11004-019-09832-6},
  url = {https://arxiv.org/abs/1806.03720}
}

@inproceedings{mosser2024explorationrobot,
  author = {Mosser, L. and Aursand, P. and Brakstad, K. S. and Lehre, C. and Myhre-Bakkevig, J.},
  title = {Exploration Robot Chat: Uncovering Decades of Exploration Knowledge and Data with Conversational Large Language Models},
  booktitle = {SPE Norway Subsurface Conference},
  year = {2024},
  doi = {10.2118/218439-MS}
}

\newpage
\appendix
\onecolumn

\section{Compact Target Schema}
\label{app:schema}

The compact target is a JSON object with optional metadata fields and a required chronological sequence:
\begin{schemajsonbox}
{"id": string, "seed": integer, "seq": [event, ...]}
\end{schemajsonbox}
Each \texttt{event} is one of:
\begin{eventjsonbox}
["D", "basement" | "layer_N"]
["T", ["basement" | "layer_N", ...], "clockwise" | "counterclockwise"]
["F", "fault_id", "normal" | "reverse"]
["E", "angular_unconformity", ["basement" | "layer_N" | "fault_id" | "intrusion_id", ...]]
["I", "intrusion_id", start, end, ["basement" | "layer_N", ...]]
\end{eventjsonbox}
Intrusion endpoints are strings: \texttt{"bottom"} or \texttt{"top"} for diagram boundaries, \texttt{"layer\_N"} for a preserved layer endpoint, and \texttt{"angular\_unconformity"} when erosion truncates the intrusion.
Every target starts with \texttt{["D","basement"]}; other fully eroded or occluded events are omitted even if they occurred in the hidden simulator history.
Basement denotes pre-existing rock below the first deposited layer.
In diagram-domain images, it is rendered as a gray/taupe unit with a rock texture made of dark diagonal crack lines and small dark speckles, and can be visible either at the bottom of the frame or where deformation exposes lower material.
All non-basement diagram colors and textures are randomized per example, with contrast constraints between neighboring depositional units, so they are not semantic layer identifiers.
For seismic-domain scoring, we accept either a leading \texttt{["D","basement"]} event or a sequence that starts directly at \texttt{["D","layer\_1"]}, because basement is not always uniquely distinguishable in seismic-style images.

\section{Model Prompts}
\label{app:prompts}

The benchmark uses a single user message containing the image and one of the following domain-specific text prompts.
The first sentence identifies the rendered image domain; the remaining instructions define the compact JSON interface, visible-evidence convention, and basement convention.

\begin{promptbox}[title={Diagram-Domain Prompt}]
You are given a synthetic sequence stratigraphic cross-section image. Infer the chronological geological event sequence. Return ONLY compact JSON. No markdown. No prose. No descriptions. Use exactly this schema: {"seq":[[CODE,...],[CODE,...]]}. Codes: D=deposition as ["D","basement"] or ["D","layer_N"]; T=tilting as ["T",[affected_layers],"clockwise|counterclockwise"]; F=faulting as ["F","fault_id","normal|reverse"]; E=erosion/unconformity as ["E","angular_unconformity",[truncated_layers_or_structures]]; I=intrusion as ["I","intrusion_id",start,end,[cut_layers]]. For intrusions, start/end are visible preserved endpoints: use "bottom" or "top" for diagram boundaries, "layer_N" for a layer endpoint, and "angular_unconformity" when the intrusion is visibly truncated by the unconformity. For erosion, list visible older layers, faults, and intrusions truncated by the unconformity. Use only D,T,F,E,I. Always start with ["D","basement"]. Include ["D","basement"] even if basement is not visible in the final image. Do not group deposition events. Emit one D event per layer. Basement is the pre-existing rock below the first deposited layer. In diagram images, visible basement is the gray/taupe unit with a rock texture made of dark diagonal crack lines and small dark speckles; it may appear at the base or in deformation-created exposures. For all non-basement events, only include events with visible evidence in the final image; omit fully eroded or occluded events.
\end{promptbox}

\begin{promptbox}[title={Seismic-Domain Prompt}]
You are given a seismic-style stratigraphic cross-section image. Infer the chronological geological event sequence. Return ONLY compact JSON. No markdown. No prose. No descriptions. Use exactly this schema: {"seq":[[CODE,...],[CODE,...]]}. Codes: D=deposition as ["D","basement"] or ["D","layer_N"]; T=tilting as ["T",[affected_layers],"clockwise|counterclockwise"]; F=faulting as ["F","fault_id","normal|reverse"]; E=erosion/unconformity as ["E","angular_unconformity",[truncated_layers_or_structures]]; I=intrusion as ["I","intrusion_id",start,end,[cut_layers]]. For intrusions, start/end are visible preserved endpoints: use "bottom" or "top" for diagram boundaries, "layer_N" for a layer endpoint, and "angular_unconformity" when the intrusion is visibly truncated by the unconformity. For erosion, list visible older layers, faults, and intrusions truncated by the unconformity. Use only D,T,F,E,I. Always start with ["D","basement"]. Include ["D","basement"] even if basement is not visible in the final image. Do not group deposition events. Emit one D event per layer. Basement is the pre-existing rock below the first deposited layer. In diagram images, visible basement is the gray/taupe unit with a rock texture made of dark diagonal crack lines and small dark speckles; it may appear at the base or in deformation-created exposures. For all non-basement events, only include events with visible evidence in the final image; omit fully eroded or occluded events.
\end{promptbox}

\section{Extended Paired Transfer Examples}
\label{app:extended-transfer-examples}

\Cref{fig:appendix-seismic-transfer-examples} shows additional held-out diagram and seismic-style observations for the same hidden scenes and compact targets.

\begin{center}
  \begin{minipage}{0.86\textwidth}
  \centering
  \includegraphics[width=\textwidth]{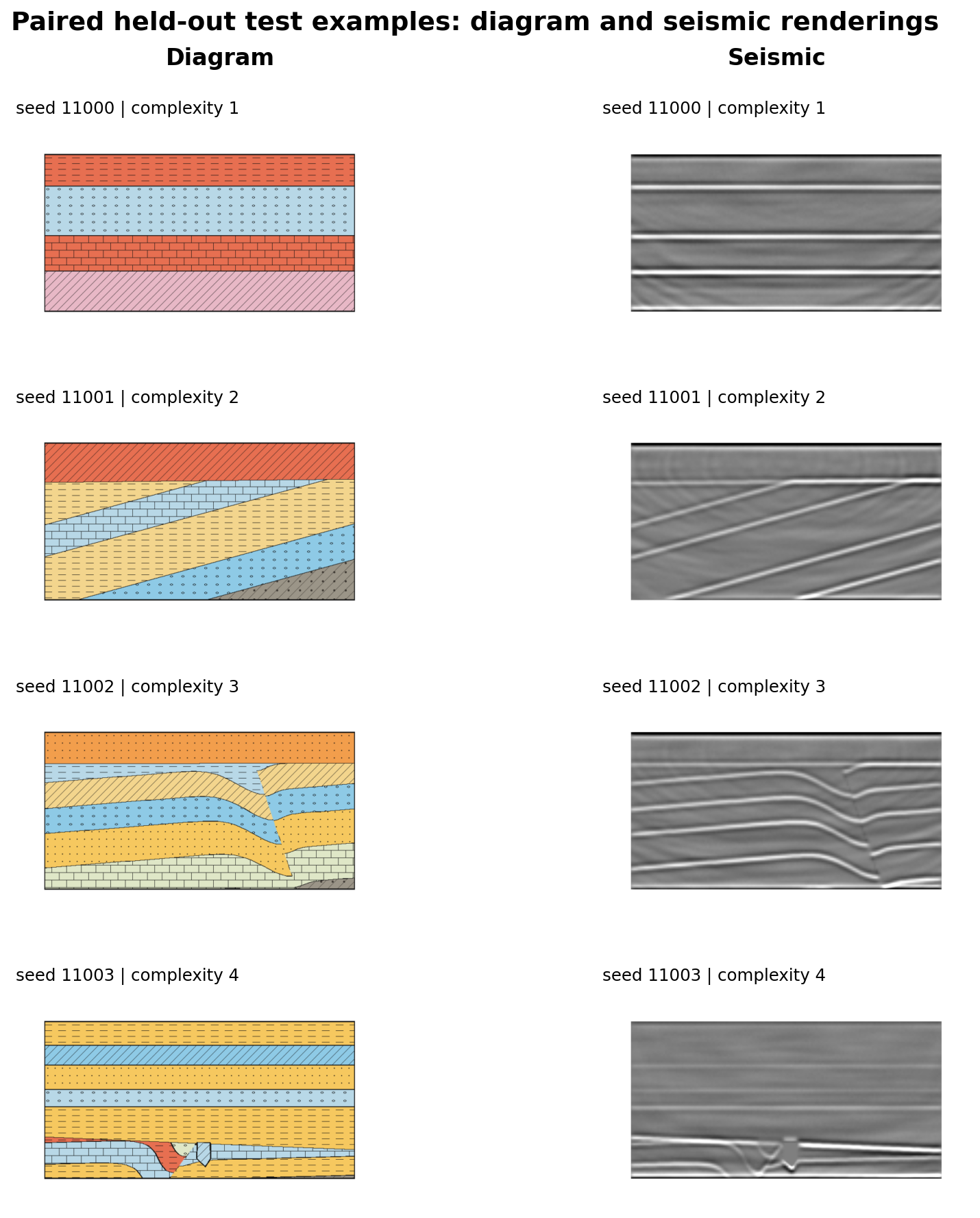}
  \captionof{figure}{Additional paired held-out transfer examples. Each row uses the same hidden scene and answer key, observed once as the diagram task and once as a seismic-style amplitude section. This pairing isolates observation-domain transfer from changes in geological history.}
  \label{fig:appendix-seismic-transfer-examples}
  \end{minipage}
\end{center}
\clearpage

\section{Generator and Renderer Specification}
\label{app:generator-spec}

This appendix records the code-level generator and renderer used for the results in this paper.
The generator is deterministic conditional on seed $s$ and complexity $c$, but it samples a geological event program before rendering.
Let $\Omega=[m,W-m]\times[m,H-m]$ be the visible frame, with $W=900$, $H=520$, and $m=70$ pixels.
The frame margin keeps contacts, faults, and intrusion endpoints away from image boundaries, leaving visible context for clipping and reducing ambiguous edge-touching geometries.
For a given seed, the simulator samples an ordered event program
\begin{align}
\mathcal{E}_{s,c}=(e_1,\ldots,e_T), \qquad
G_t = A_{e_t}(G_{t-1};\theta_t),
\end{align}
where $G_t$ is the geometric scene state after event $t$, $A_{e_t}$ is the event operator, and $\theta_t$ are event parameters.
The final visible target is derived from $G_T$ by retaining only events with visible evidence in the rendered scene.

\begin{center}
\resizebox{\textwidth}{!}{%
\begin{tikzpicture}[node distance=4mm and 4mm]
  \node[pipelinebox,text width=0.13\textwidth] (seed) {seed $s$\\complexity $c$};
  \node[pipelinebox,text width=0.14\textwidth,right=of seed] (program) {sample event\\program $\mathcal{E}_{s,c}$};
  \node[pipelinebox,text width=0.15\textwidth,right=of program] (ops) {apply geological\\operators $A_e$};
  \node[pipelinebox,text width=0.14\textwidth,right=of ops] (scene) {clip and scale\\final geometry $G_T$};
  \node[pipelinebox,text width=0.13\textwidth,right=of scene] (target) {visible-event\\JSON target};
  \node[pipelinebox,text width=0.13\textwidth,right=of target] (images) {diagram or\\seismic image};
  \draw[pipelinearrow] (seed) -- (program);
  \draw[pipelinearrow] (program) -- (ops);
  \draw[pipelinearrow] (ops) -- (scene);
  \draw[pipelinearrow] (scene) -- (target);
  \draw[pipelinearrow] (target) -- (images);
\end{tikzpicture}
}
\end{center}

\subsection{Stratigraphic Event Program}
\label{app:strat-generator}

The default random generator samples $N\sim U\{4,\ldots,9\}$ total depositional layers.
A lower package size $N_{\mathrm{pre}}$ is sampled uniformly from
\begin{align}
N_{\mathrm{pre}} \sim U\left\{\max(2,\lfloor N/2\rfloor),\ldots,\max(2,N-1)\right\}.
\end{align}
The event order is constrained to remain geologically plausible: lower deposition, optional tilting, optional faults and intrusion, erosion, upper deposition, and for the highest complexity optional later faulting, erosion, and additional deposition.
Depositional lithology is sampled from sandstone, shale, limestone, conglomerate, and siltstone.
The numerical constants in \cref{tab:appendix-strat-params,tab:appendix-geometry-params} are synthetic generator defaults chosen to keep contacts, offsets, intrusions, and unconformities visible at the benchmark image resolution.
They are not calibrated field or restoration parameters.

\begin{table}[h]
\centering
\small
\caption{Stratigraphic event-program sampling used by \texttt{generate\_random\_scenario}. Uniform ranges are inclusive for integer variables and continuous for real-valued variables.}
\label{tab:appendix-strat-params}
\renewcommand{\arraystretch}{1.15}
\begin{tabular}{p{0.22\textwidth}p{0.23\textwidth}p{0.46\textwidth}}
\toprule
Quantity & Range or rule & Role \\
\midrule
Total layers $N$ & $U\{4,\ldots,9\}$ & Number of depositional units in the base program. \\
Lower package $N_{\mathrm{pre}}$ & $U\{L_N,\ldots,N-1\}$, $L_N=\max(2,\lfloor N/2\rfloor)$ & Layers deposited before deformation and erosion. \\
Layer thickness $h_i$ & $U\{34,\ldots,62\}$ px & Depositional-unit thickness before later deformation. \\
Tilt angle $\alpha$ & $U(-10^\circ,10^\circ)$, active if $c\ge2$ with probability $0.75$ & Rotates the existing lower sequence. \\
Fault count & one if $3\le c<5$; $U\{1,2,3\}$ if $c=5$ & Number of early finite faults. \\
Fault center $x_f$ & $U(0.28,0.72)$ of frame width & Horizontal fault anchor. \\
Fault throw $\tau$ & $\pm U(45,95)$ px & Vertical throw magnitude and sense. \\
Fault dip $\phi$ & $U(44^\circ,72^\circ)$ & Fault-trace inclination. \\
Fault half length $L$ & $U(280,430)$ px & Finite along-fault support. \\
Damage width $D$ & $U(62,95)$ px & Across-fault attenuation scale. \\
Core width $w_c$ & $U(2.5,5.5)$ px & Side disambiguation near the fault core. \\
Intrusion probability & $0.45$ before erosion for $c\ge4$; guaranteed later if absent & Ensures complex examples include igneous cross-cutting evidence. \\
Intrusion width & $U(24,48)$ px & Dike width. \\
Intrusion top/bottom & top $U(0.08,0.45)H$, bottom $U(0.70,0.97)H$ & Vertical endpoints before clipping. \\
Erosion relief and slope & relief $U(-22,26)$ px, slope $U(-0.04,0.04)$ & Defines the angular-unconformity surface. \\
Late complexity branch & probability $0.4$ if $c=5$ & Adds a younger fault, second erosion, and $U\{1,2\}$ layers. \\
\bottomrule
\end{tabular}
\end{table}

\subsection{Geometric Operators}
\label{app:geometry-operators}

The initial basement is a polygon below the first depositional surface and is rendered with the fixed gray/taupe crack-and-speckle basement texture.
Deposition uses the current top surface $s_{t-1}(x)$ to create a polygonal layer.
With contact overlap $\delta=2$ px and thickness $h_i$, the nominal top and bottom surfaces are
\begin{align}
b_i(x) &= s_{t-1}(x)+\delta,\\
u_i(x) &= \max(m, b_i(x)-h_i).
\end{align}
If the current surface is an unconformity, the new layer top is horizontal at $\min_x b_i(x)-h_i$; otherwise it follows the prior top surface.
When the scene approaches the top frame, all previous geometry is translated downward to leave at least $24$ px of visible thickness for the next deposition.

Tilting rotates all existing layer polygons around the lower-frame pivot $o=(W/2,H-m)$:
\begin{align}
P_i' = R_\alpha(P_i;o).
\end{align}
Faulting uses a finite displacement field in local fault coordinates.
For a point $p=(x,y)$, center $q=(x_f,0.55H)$, fault tangent
\begin{align}
\mathbf{t}=(d\cos\phi,\sin\phi), \qquad \mathbf{n}=(-t_y,t_x),
\end{align}
and local coordinates $a=(p-q)\cdot\mathbf{t}$, $b=(p-q)\cdot\mathbf{n}$, the displacement amplitude is
\begin{align}
\lambda(a,b)
&=\sqrt{\max\left(0,1-\left(a/L\right)^2\right)}
  \exp\left[-\left(|b|/D\right)^2\right] \notag\\
&\quad\cdot
\max\left(\mathbf{1}[m_f b>0],\,
0.18\exp\left[-\left(b/(0.45D)\right)^2\right]\right),
\end{align}
where $m_f\in\{-1,1\}$ is the moving-side sign.
The slip vector is $\mathbf{v}=(\tau/t_y)\mathbf{t}$, so the deformed point is
\begin{align}
p' = p + \lambda(a,b)\mathbf{v}.
\end{align}
Layer and intrusion polygons are split by the fault trace before this deformation is applied.
If a younger fault terminates against an older fault, the trace is truncated at their intersection and deformation is applied only on the active side.

Erosion constructs a linear unconformity
\begin{align}
u_e(x)=y_0+\beta(x-W/2),
\end{align}
where $\beta$ is the sampled slope and $y_0$ is the largest value required to keep the surface at least $8$ px below the visible top of the eroded package.
All material above $u_e(x)$ is clipped.
Intrusion inserts a six-vertex dike polygon with sampled center, width, top, and bottom; later erosion and faulting can clip or deform this polygon like any other geometry.
Finally, all visible geometry is vertically scaled into the frame if the constructed scene does not fill the available height.
Remaining visible voids are made explicit: voids connected to the top frame extend the youngest visible layer upward, while lower voids become basement exposure.

\begin{table}[h]
\centering
\small
\caption{Main geometric constants and operators in the stratigraphic generator.}
\label{tab:appendix-geometry-params}
\renewcommand{\arraystretch}{1.15}
\begin{tabular}{p{0.22\textwidth}p{0.24\textwidth}p{0.45\textwidth}}
\toprule
Symbol or setting & Value & Meaning \\
\midrule
$W,H$ & $900,520$ px & Native scene width and height. \\
$m$ & $70$ px & Visible frame margin for boundary context and stable clipping. \\
$\delta$ & $2$ px & Contact overlap used to avoid raster gaps. \\
Minimum visible thickness & $24$ px & Downward translation threshold before a new layer is deposited. \\
Fault sample spacing & $12$ px & Polygon-edge densification before fault deformation. \\
Fault trace length & $1.28L$ on each side of center & Displayed and splitting trace for a finite fault. \\
Erosion clearance & $8$ px below visible top & Ensures erosion cuts existing material rather than floating above it. \\
Final vertical scaling & bottom-fixed affine scale & Expands constructed material into the visible frame. \\
Void fill convention & explicit basement or top-layer extension & Prevents deformation gaps from being rendered as unlabeled geology. \\
\bottomrule
\end{tabular}
\end{table}

\subsection{Seismic-Domain Synthetic Renderer}
\label{app:seismic-generator}

The seismic-style renderer consumes the same final scene $G_T$ and answer key as the diagram renderer.
It does not resample the geology.
Instead, it maps the final geometry to acoustic material properties, simulates simple surface seismic acquisition with a two-dimensional acoustic wave equation, images the simulated records back into depth, and displays the result as a grayscale amplitude section.
The goal is not survey realism; it is a controlled observation domain in which the same layers, faults, unconformities, and intrusions are visible through seismic-style amplitudes.

\begin{center}
\resizebox{\textwidth}{!}{%
\begin{tikzpicture}[node distance=4mm and 4mm]
  \node[pipelinebox,text width=0.13\textwidth] (geom) {final\\geometry};
  \node[pipelinebox,text width=0.13\textwidth,right=of geom] (props) {rasterize\\$v_p,\rho,Z$};
  \node[pipelinebox,text width=0.13\textwidth,right=of props] (fd) {surface shots\\acoustic finite\\difference (FD)};
  \node[pipelinebox,text width=0.13\textwidth,right=of fd] (mig) {depth\\imaging};
  \node[pipelinebox,text width=0.13\textwidth,right=of mig] (blend) {contact image\\and cleanup};
  \node[pipelinebox,text width=0.13\textwidth,right=of blend] (noise) {noise, gain,\\display scaling};
  \draw[pipelinearrow] (geom) -- (props);
  \draw[pipelinearrow] (props) -- (fd);
  \draw[pipelinearrow] (fd) -- (mig);
  \draw[pipelinearrow] (mig) -- (blend);
  \draw[pipelinearrow] (blend) -- (noise);
\end{tikzpicture}
}
\end{center}

Let $M_k(x,y)$ be the raster mask for deposited unit $k$ in the final deformed scene.
For each unit, we assign density
\begin{align}
\rho_k=\operatorname{clip}\left(1980+36k+d_{\ell_k},1850,2850\right),
\end{align}
where $d_{\ell_k}$ is a small lithology-dependent density offset.
We then assign acoustic impedance
\begin{align}
Z_k = 3.8 + 0.95k + 0.08\,\ell_k,
\end{align}
where $\ell_k$ is a bounded lithology offset.
The fixed $0.95$ step guarantees that every successive deposited layer differs in acoustic impedance enough to produce a visible contact reflection.
The P-wave velocity is derived from the impedance and density,
\begin{align}
v_{p,k}=\operatorname{clip}\left(\frac{10^6 Z_k}{\rho_k},1750,4300\right),
\end{align}
where the factor $10^6$ converts the compact impedance units used in the renderer to International System of Units (SI)-like units.
The raster fields are
\begin{align}
v_p(x,y)=\sum_k M_k(x,y)v_{p,k},\qquad
\rho(x,y)=\sum_k M_k(x,y)\rho_k,\qquad
Z(x,y)=\rho(x,y)v_p(x,y)/10^6.
\end{align}
Intrusions are assigned a high-impedance body with $v_p=4300$ and $\rho=2920$.
After imaging, amplitudes inside the intrusion mask are smoothed and attenuated so that intrusions appear as coherent bodies that cut layers rather than as strings of point-diffraction artifacts.

The velocity raster is resized to a compact modeling grid and padded by $32$ cells on all sides.
We use a second-order acoustic finite-difference update for pressure $p$,
\begin{align}
p^{t+1}_{i,j}
=2p^t_{i,j}-p^{t-1}_{i,j}
{}+\left(\frac{v_p(i,j)\Delta t}{\Delta x}\right)^2\nabla^2 p^t_{i,j}
{}+s^t_{i,j},
\end{align}
where $\nabla^2$ is the five-point spatial Laplacian, $\Delta x=12$ m, $\Delta t=0.001$ s, and $s^t$ is a Ricker source wavelet.
A multiplicative sponge taper in the padded region absorbs outgoing energy and keeps boundary reflections outside the displayed domain.
The default acquisition uses $11$ surface shots and $88$ surface receivers, both placed just below the top boundary.
For each shot, recorded traces are direct-arrival muted using the near-surface velocity and then de-meaned across receivers and time to reduce source and receiver imprint.

The simulated shot gathers are imaged back to depth by a lightweight Kirchhoff-style condition.
For image point $(x,z)$, shot position $x_s$, receiver position $x_r$, and depth-dependent root-mean-square (RMS) velocity $v_{\mathrm{rms}}(z)$, we use the two-way travel-time approximation
\begin{align}
T(x,z;x_s,x_r)
=\frac{\sqrt{(x-x_s)^2+z^2}+\sqrt{(x-x_r)^2+z^2}}{v_{\mathrm{rms}}(z)}.
\end{align}
The image is the illumination-normalized weighted sum of trace amplitudes sampled at $T$:
\begin{align}
I_{\mathrm{mig}}(x,z)
=
\frac{
\sum_{s,r} D_{s,r}\!\left(T(x,z;x_s,x_r)\right)\,a(x,z;x_s,x_r)
}{
\sum_{s,r} a(x,z;x_s,x_r)+10^{-6}
},
\end{align}
where $D_{s,r}$ is a modeled shot gather and $a$ is a geometric spreading weight.
The very shallow part of the image is tapered to suppress source and receiver artifacts.
The broad migration background is removed by subtracting a smoothed copy, leaving a wave-propagation texture term.

To keep the section interpretable as a depth-domain observation, the displayed image is dominated by the impedance-contact image.
We compute vertical reflectivity from $Z$,
\begin{align}
R_Z(y,x)=\frac{Z(y+1,x)-Z(y,x)}{\max(Z(y+1,x)+Z(y,x),10^{-6})},
\end{align}
convolve it vertically with a short Ricker wavelet, normalize it, and blend it with the high-pass migration texture:
\begin{align}
S_0=0.12\,\operatorname{norm}(I_{\mathrm{mig}}-\mathcal{G}_{6,10}*I_{\mathrm{mig}})
+0.88\,\operatorname{norm}(w *_y R_Z).
\end{align}
Here $*_y$ denotes vertical convolution.
The migration texture is suppressed in a halo around intrusions, and the intrusion body is smoothed after blending and again after display-grid resizing.
Finally, trace-gain variation, coherent noise, and white noise are applied:
\begin{align}
\tilde{S}(x,y)
&= S_0(x,y)\left(1+\sigma_g \tilde{g}(x)\right)
 + \sigma_c\,\mathrm{std}(S_0)\,\tilde{C}(x,y) \notag\\
&\quad + \epsilon(x,y), \qquad
\epsilon\sim\mathcal{N}\left(0,\sigma_n^2\mathrm{std}(S_0)^2\right),
\end{align}
where $\tilde{g}$ and $\tilde{C}$ are unit-standard-deviation smoothed random fields.
The grayscale image intensity is
\begin{align}
I(x,y)=\operatorname{clip}\left(127.5\left[1-\tilde{S}(x,y)/Q_{98.5}(|\tilde{S}|)\right],0,255\right).
\end{align}
The numerical constants in \cref{tab:appendix-seismic-params} are synthetic renderer defaults chosen to keep contact amplitudes, offsets, intrusions, and unconformities visible at the benchmark image resolution.
They are not calibrated field-seismic or rock-physics parameters.

\begin{table}[h]
\centering
\small
\caption{Seismic-domain renderer parameters. These values are the defaults in \texttt{SeismicRenderConfig}.}
\label{tab:appendix-seismic-params}
\renewcommand{\arraystretch}{1.15}
\begin{tabular}{p{0.27\textwidth}p{0.18\textwidth}p{0.46\textwidth}}
\toprule
Parameter & Value & Role \\
\midrule
Model grid & $176\times104$ cells & Acoustic finite-difference grid before resizing to the image domain. \\
Boundary padding & $32$ cells & Absorbing sponge thickness around the modeled domain. \\
Shot count & $11$ & Surface source positions used for illumination. \\
Receiver count & $88$ & Surface receiver positions per shot. \\
Grid spacing $\Delta x$ & $12$ m & Spatial sampling for the acoustic model. \\
Time step $\Delta t$ & $0.001$ s & Temporal sampling for wave propagation. \\
Record length & $1.15$ s & Modeled shot-gather duration. \\
Source frequency & $24$ Hz & Ricker source peak frequency. \\
Direct mute & $0.16$ s & Extra direct-arrival mute after the source--receiver travel time. \\
Base impedance & $3.8$ & Compact acoustic-impedance value for the pre-deposition baseline. \\
Layer impedance step & $0.95$ & Guaranteed minimum trend separating successive deposited units. \\
Lithology impedance scale & $0.08$ & Small lithology perturbation added to the layer trend. \\
Structural blend & $0.88$ & Weight of the impedance-contact image relative to migration texture. \\
White-noise scale $\sigma_n$ & $0.025$ & Independent noise as a fraction of signal standard deviation. \\
Coherent-noise scale $\sigma_c$ & $0.09$ & Smoothed band-limited noise amplitude. \\
Coherent-noise blur & $\sigma_y=1.0$, $\sigma_x=7.0$ px & Vertical and lateral smoothing for coherent noise. \\
Trace-gain scale $\sigma_g$ & $0.05$ & Lateral trace-amplitude variation. \\
Trace-gain smoothing & $28.0$ px & Lateral Gaussian smoothing for trace gain. \\
Display percentile & $98.5$ & Robust amplitude normalization. \\
Intrusion texture suppression & $0.95$ & Suppresses migration texture in and around high-contrast intrusion bodies. \\
Intrusion smoothing & $\sigma_y=8.0$, $\sigma_x=2.5$ px & Smooths the final amplitude inside intrusion masks. \\
Intrusion amplitude scale & $0.05$ & Residual amplitude retained inside smoothed intrusion bodies. \\
Colormap & gray & Display mode used for paper and evaluation. \\
\bottomrule
\end{tabular}
\end{table}

\clearpage
\section{Artifacts and Benchmark Details}
\label{app:benchmark-details}

The repository, fixed dataset identifiers, model references, and LoRA adapter identifiers used for this paper are collected here; anonymous review access is provided for non-public artifacts.

\begin{table}[h]
\centering
\small
\caption{Repository and dataset artifacts.}
\label{tab:appendix-artifact-links}
\renewcommand{\arraystretch}{1.15}
\begin{tabular}{p{0.20\textwidth}p{0.46\textwidth}p{0.26\textwidth}}
\toprule
Artifact & Link & Role \\
\midrule
Code repository & \url{https://github.com/LukasMosser/geo-strat-rl} & Source code, notebooks, figure scripts, and paper assets. \\
Training dataset & \url{https://huggingface.co/datasets/porestar/geo-strat-rl-train} & Seeds 50000--66383 for fixed cached training examples. \\
Validation dataset & \url{https://huggingface.co/datasets/porestar/geo-strat-rl-val} & Seeds 10000--10099 for prompt, reward, and checkpoint selection. \\
Held-out test dataset & \url{https://huggingface.co/datasets/porestar/geo-strat-rl-test} & Seeds 11000--11099 for final reporting. \\
\bottomrule
\end{tabular}
\end{table}

\begin{table}[h]
\centering
\small
\caption{Model and adapter artifacts.}
\label{tab:appendix-model-links}
\renewcommand{\arraystretch}{1.15}
\begin{tabular}{p{0.22\textwidth}p{0.42\textwidth}p{0.28\textwidth}}
\toprule
Artifact & Link or identifier & Role \\
\midrule
Qwen2.5-VL-3B base & \url{https://huggingface.co/Qwen/Qwen2.5-VL-3B-Instruct} & Local pretrained baseline and RLVR base model. \\
Qwen3-VL-4B base & \url{https://huggingface.co/Qwen/Qwen3-VL-4B-Instruct} & Local pretrained baseline and RLVR base model. \\
Qwen2.5-VL-3B diagram ng16 LoRA & \url{https://huggingface.co/porestar/qwen25vl3b_g4_seed50000_ng16-lora} & Diagram-domain adapter used for final diagram and transfer evaluation. \\
Qwen3-VL-4B diagram ng16 LoRA & \url{https://huggingface.co/porestar/qwen3vl4b_g4_seed80000_ng16-lora} & Diagram-domain adapter used for final diagram and transfer evaluation. \\
Qwen3-VL-4B seismic ng16 LoRA & \url{https://huggingface.co/porestar/qwen3vl4b_g4_seed80000_seismic_ng16-lora} & Seismic-domain adapter used for reverse transfer evaluation. \\
Gemma 4 31B IT & \url{https://openrouter.ai/google/gemma-4-31b-it} & Pretrained baseline. \\
Qwen3-VL-30B-A3B & \url{https://openrouter.ai/qwen/qwen3-vl-30b-a3b-instruct} & Pretrained baseline. \\
Mistral Small 3.2 24B & \url{https://openrouter.ai/mistralai/mistral-small-3.2-24b-instruct} & Pretrained baseline with a 100-example, 5-rollout confidence interval. \\
Gemini 2.5 Flash Lite & \url{https://openrouter.ai/google/gemini-2.5-flash-lite} & Pretrained baseline. \\
\bottomrule
\end{tabular}
\end{table}

The current held-out test split is \texttt{porestar/geo-strat-rl-test}, split \texttt{train}, containing 100 examples with seeds 11000--11099.
The pretrained-model notes in \texttt{docs/benchmarks/} record API-service run context, and the machine-checkable artifact manifest records the canonical API uncertainty summaries used for the main OpenRouter rows.
The GPT-5.5 diagram comparison is backed by the Azure OpenAI 100-example, 3-rollout summary; the GPT-5.5 seismic comparison is backed by the Azure OpenAI 100-example, 5-rollout seismic summary.
The canonical local base and RLVR evaluation artifacts are bundled in \texttt{docs/assets/eval\_loras\_15\_06\_2026/batched\_geo\_strat\_evals.zip}; it contains per-rollout JSON Lines (JSONL) rows with prompts, targets, generated responses, parsed outputs, reward components, bootstrap summaries, and uncertainty files for the 100-example, 5-rollout diagram and seismic-style runs.
Bootstrap confidence intervals for any evaluation JSONL can be computed with \texttt{geo-strat-rl-eval-uncertainty}.
The interval calculation groups rows by example, averages all rollouts for each example, and bootstraps over example-level means; final intervals should be recomputed after target-schema refreshes rather than copied across schema versions.
The reported GRPO training curves are tracked in Weights \& Biases and rendered in \cref{fig:grpo-reward-overlay}.
The reproduction notebook is \texttt{notebooks/colab\_batched\_post\_rl\_eval.ipynb}; it evaluates the Qwen2.5-VL-3B and Qwen3-VL-4B base checkpoints, the diagram ng16 LoRA adapters, and the Qwen3-VL-4B seismic ng16 LoRA adapter on both the diagram and seismic-style domains.
All paper figure assets are regenerated by \texttt{scripts/generate\_paper\_figures.py}; the optional \texttt{--refresh-wandb-history} flag refreshes the cached W\&B reward history used for \cref{fig:grpo-reward-overlay}.

\end{document}